\documentclass[journal,twoside,web]{ieeecolor}
\usepackage{generic}
\usepackage{cite}
\usepackage{amsmath,amssymb,amsfonts}
\usepackage{algorithm}
\usepackage{algorithmic}
\usepackage{graphicx}
\usepackage{textcomp}
\usepackage{mathtools}
\usepackage{booktabs,siunitx}
\usepackage{float}
\usepackage{placeins}
\newcommand{\mc}[1]{\multicolumn{1}{@{}c@{}}{#1}}

\newcommand{\MR}{\mbox{\textit{MR}}}
\newcommand{\IAE}{\mbox{\textit{IAE}}}

\def\BibTeX{{\rm B\kern-.05em{\sc i\kern-.025em b}\kern-.08em
    T\kern-.1667em\lower.7ex\hbox{E}\kern-.125emX}}
\begin{document}
\title{A Reinforcement Learning Approach for Transient Control of Liquid Rocket Engines}
\author{G\"unther Waxenegger-Wilfing$^*$, Kai Dresia$^*$, Jan Deeken, and Michael Oschwald
\thanks{$^*$ Both authors contributed equally to this work.}
\thanks{The authors are with the Department of Rocket Propulsion, German Aerospace Center (DLR) Institute of Space Propulsion, Hardthausen am Kocher, Germany (e-mail: guenther.waxenegger@dlr.de, kai.dresia@dlr.de, jan.deeken@dlr.de, michael.oschwald@dlr.de).}
\thanks{\copyright 2020 IEEE. Personal use of this material is permitted. Permission from IEEE must be obtained for all other uses, in any current or future media, including reprinting/republishing this material for advertising or promotional purposes, creating new collective works, for resale or redistribution to servers or lists, or reuse of any copyrighted component of this work in other works.}}


\maketitle

\begin{abstract}
Nowadays, liquid rocket engines use closed-loop control at most near steady operating conditions. The control of the transient phases is traditionally performed in open-loop due to highly nonlinear system dynamics. This situation is unsatisfactory, in particular for reusable engines. The open-loop control system cannot provide optimal engine performance due to external disturbances or the degeneration of engine components over time. In this paper, we study a deep reinforcement learning approach for optimal control of a generic gas-generator engine's continuous start-up phase. It is shown that the learned policy can reach different steady-state operating points and convincingly adapt to changing system parameters. A quantitative comparison with carefully tuned open-loop sequences and PID controllers is included. The deep reinforcement learning controller achieves the highest performance and requires only minimal computational effort to calculate the control action, which is a big advantage over approaches that require online optimization, such as model predictive control.
\end{abstract}


\section{Introduction}
\label{sec:introduction}
\IEEEPARstart{T}{HE} demands on the control system of liquid rocket engines have significantly increased in recent years \cite{colas2019}, in particular for reusable engines. Advanced mission scenarios, e.g. in-orbit maneuvers or propulsive landings, require deep throttling and re-start capabilities. The aging of reusable engines also requires a robust control system as the performance of engine components might degrade over time, e.g. due to soot depositions \cite{meland1989, lausten1985, bossard1989}, increased leakage mass flows caused by seal aging \cite{roelke1973}, or turbine blade erosions \cite{hampson1984}.
The cost-efficient operation of a reusable launch vehicle is only possible if the engines possess a long service life without expensive maintenance.

Nowadays, most liquid rocket engines use predefined valve sequences to drive the system from the start signal to a desired steady-state and to shut down the engine safely. These control sequences are usually determined during costly ground tests. Closed-loop control is at most used near steady operating conditions to maintain a desired combustion chamber pressure and mixture ratio \cite{lorenzo1992}. The resulting lower deviations of the controlled variables decrease the amount of extra propellant to be carried, which in turn increases the payload capacity of the launch vehicle. Although the importance of closed-loop control has been evident for many years, the majority of rocket engines still employ valves which are operated with pneumatic actuators, too inefficient for a sophisticated closed-loop control system. The development of an all-electric control system started in the late 90s in Europe \cite{chopinet2012}. The future European Prometheus engine will have such a system \cite{iannetti2017}.
Other countries are also well advanced in the research and development of electrically operated flow control valves \cite{asakawa2019}. Due to the electrification of the actuators and the grown demands, the interest in closed-loop solutions increased recently and will further rise in the future.

Furthermore, optimal control of the engine operation, including the transient phases, is the only way to realize high performing systems, which also comply with the aforementioned demands on the control system of future liquid rocket engines \cite{perez-roca2019}. One way to solve optimal control problems is to use reinforcement learning (RL). Although the application of such modern methods of artificial intelligence seems unorthodox in this setting, it offers certain advantages. First, given a suitable simulation environment, RL algorithms can automatically generate optimal transient sequences. Second, the trained RL controller features a minimal computational effort to calculate the control action, so it can easily be used for closed-loop control of the demanding transient phases. Third, RL is perfectly suited for complex control tasks, including multiple objectives and multiple regimes \cite{lopez2019}. Optimal control using RL \cite{kiumarsi2018} has been studied in many different areas, from robotics \cite{gu2016, yang2018} and medical science \cite{mahmud2018} to flight control \cite{heyer2020, gaudet2020} and process control \cite{spielberg2017}. Furthermore, the benefits of an intelligent engine control system, where artificial intelligence techniques are used for control reconfiguration and condition monitoring, have already been investigated in the space shuttle area \cite{musgrave1992a, nemeth1991}. 

The objective of our work is analogous to the investigation of P\'erez-Roca et al. \cite{perez-roca2019b}, where a model predictive control (MPC) approach to control the start-up transient of a liquid rocket engine was studied. After the derivation of a suitable state-space model \cite{perez-roca2018}, a linear MPC controller was synthesized. The controller completes the start-up and can track the end-state references with sufficient accuracy. MPC and RL have specific advantages and disadvantages. The work presented here aims to evaluate the capabilities and limitations of RL for liquid rocket engine control.

\noindent Our main contributions are the following:
\begin{itemize}
    \item formulation of optimal start-up control as a RL problem
    \item training and evaluation of the RL controller for multiple operating conditions and degrading turbine efficiencies
    \item quantitative comparison with carefully tuned open-loop sequences and PID controllers 
\end{itemize}

\noindent The remainder of this paper is structured as follows: Section \ref{sec:rl} describes the basics of RL and presents pseudocode of the used RL algorithm. The simulation environment and its coupling with the RL algorithms are outlined in section \ref{sec:sim_environment}. Section \ref{sec:test_case} discusses the test case. Section \ref{sec:results} reports the results, including the comparison with the performance of PID controllers. Finally, section \ref{sec:conclusion} provides concluding remarks.

\section{Reinforcement Learning}
\label{sec:rl}
In this section, we review basic RL concepts \cite{sutton2018}. RL algorithms can be used to solve optimal control problems stated as Markov decision processes (MDPs) \cite{bertsekas2019}. MDPs provide a mathematical framework for modeling decision making in situations where the system changes possibly in a stochastic manner. Standard MDPs work in discrete time: at each time step, the controller (usually called the agent in RL) receives information on the state of the system and takes an action in response. The decision rule is called a policy in RL. The action changes the state of the system, and the latest transition is evaluated via a reward function. The optimal control objective is to maximize the (expected) cumulative reward from each initial state. Formally, an MDP consists of the state-space $X$ of the system, the action (input) space $U$, the transition function (dynamics) $f$ of the system, and the reward function $\rho$ (negative costs). Due to the origins of the field in artificial intelligence, the usual notation would be $S$ for the state-space, $A$ for the action space, $P$ for the dynamics, and $R$ for the reward function. In this paper, notation inspired by control theory is used. As a result of the action $u_k$ applied in state $x_k$ at discrete time step $k$, the state changes to $x_{k+1}$ and a scalar reward $r_{k+1}=\rho(x_k,u_k,x_{k+1})$ is received. The goal is to find a policy $\pi$, so that $u_k=\pi(x_k)$, that maximizes the cumulative reward, typically the expected discounted sum over the infinite horizon:
\begin{equation}
    \mathbb{E}_{x_{k+1}\sim f(x_k,\pi(x_k),\cdot)}\Bigg\{\sum_{k=0}^{\infty}\gamma^k\rho(x_k,\pi(x_k),x_{k+1})\Bigg\},
\end{equation}
where $\gamma\in (0,1]$ is the discount factor. The mapping from a state $x_0$ to the value of the cumulative reward for a policy $\pi$ is called the (state) value function $V^{\pi}(x_0)$:
\begin{equation}
    V^{\pi}(x_0) = \mathbb{E}_{x_{k+1}\sim f(x_k,\pi(x_k),\cdot)}\Bigg\{\sum_{k=0}^{\infty}\gamma^k\rho(x_k,\pi(x_k),x_{k+1})\Bigg\}.
\end{equation} 
The control objective is to find an optimal policy $\pi^*$ that leads to the maximal value function, for all $x_0$:
\begin{equation}
    V^*(x_0):=\max_{\pi}V^{\pi}(x_0),\forall x_0
\end{equation}
Although state-values functions suffice to define optimality, it is useful to define action-value functions, called Q-functions. The action-value function gives the expected reward if one starts in state $x$, takes an arbitrary action $u$ (which may not have come from the policy), and then forever after acts according to policy $\pi$: 
\begin{equation}
    Q^{\pi}(x,u) = \mathbb{E}_{x'\sim f(x,u,\cdot)}\{\rho(x,u,x')+\gamma V^{\pi}(x')\},
\end{equation}
where the prime notation indicates quantities at the next discrete time step. The optimal Q-function $Q^*$ is defined using $V^*$. Once an optimal Q-function $Q^*$ is available, an optimal policy $\pi^*$ can be computed by
\begin{equation}
    \pi^*(x)\in\underset{u}{\arg\max}\, Q^*(x,u),
\end{equation}
while the formula to compute $\pi^*$ from $V^*$ is more complicated. As a consequence of the definitions, the Q-functions $Q^{\pi}$ and $Q^*$ fulfill the Bellman equations:
\begin{equation}
    Q^{\pi}(x,u) = \mathbb{E}_{x'\sim f(x,u,\cdot)}\{\rho(x,u,x')+\gamma Q^{\pi}(x',\pi(x'))\}
\end{equation}
and
\begin{equation}
    Q^*(x,u) = \mathbb{E}_{x'\sim f(x,u,\cdot)}\{\rho(x,u,x')+\gamma\max_{u'}\,Q^*(x',u')\},
\end{equation}
which are of central importance in RL. The crucial advantage of RL algorithms is that they do not require a model of the system dynamics. Instead, an optimal policy can be found by learning from samples of transitions and rewards. The problem formulation with MDPs and the associated solution techniques also handle nonlinear, stochastic dynamics, and nonquadratic reward functions. Perhaps the most popular RL algorithm is Q-learning. In Q-learning, one starts from an arbitrary initial Q-function $Q_0$ and updates it using observed state transitions and rewards. The update rule is of the following form:
\begin{multline}
Q_{k+1}(x_k,u_k) = Q_k(x_k,u_k)\\
+\alpha_k[r_{k+1}+\gamma\max_{u'}\,Q_k(x_{k+1},u')-Q_k(x_k,u_k)],
\end{multline}
where $\alpha_k\in (0,1]$ is the learning rate. The term inside the square bracket is nothing else than the difference between the updated estimate of the optimal Q-value of $(x_k,u_k)$ and the current estimate $Q_k(x_k,u_k)$. Under mild assumptions on the learning rate and that a suitable exploratory policy is used to obtain samples, i.e. data tuples of the form $(x_k,u_k,x_{k+a},r_{k+1})$, Q-learning asymptotically converges to $Q^*$, which satisfies the Bellman optimality equation. The reader is referred to \cite{busoniu2018} for a description of similar RL algorithms. Q-learning and its many variants require that Q-functions and policies are exactly represented, e.g. as a table indexed by the discrete states and actions. Especially for the control of physical systems, the states and actions are continuous; moreover, exact representations are in general impossible. Normal Q-learning does not work in this setting. Fortunately, methods like Q-learning can be combined with function approximation. We denote approximate versions of the Q-function and the policy by $\hat{Q}(x,u;\theta)$ and $\hat{\pi}(x;w)$, where $\theta$ and $w$ are the parameters of parametric approximators. There are many different function approximators to choose from.

The combination of RL with deep neural networks (DNNs) as function approximators leads to the field of deep RL. In the last years, deep RL algorithms have achieved impressive results, such as reaching super-human performance in the game of Go. Besides the sensational results in board games or video games, those algorithms are successfully used in areas like robotics. In deep Q-learning, one uses a neural network to approximate the Q-function. Neural networks can represent any smooth function arbitrarily well given enough parameters, and therefore they can learn complex Q-functions. Loss functions and gradient descent optimization are used to fit the parameters of the models. Gradient estimates are usually averaged over individual gradients computed for a batch of experiences.

Nevertheless, the simple training procedure is unstable, because sequential observations are correlated, and techniques like experience replay have to be used. Correlated experiences are saved into a replay buffer. When batches of experiences are needed for training, these batches are generated by sampling from the replay buffer in a randomized order. A further reason for the simple training procedure's instability is that the target values depend on the parameters one wants to optimize. The solution is to use a so-called target network, $\hat{Q}(x,u;\theta^{-})$, with target parameters $\theta^{-}$, which slowly track the online parameters. While deep Q-learning solves problems with continuous state-spaces, it can only handle discrete and low-dimensional action spaces. The reason for that is the following: (deep) Q-learning requires fast maximization of Q-functions over actions. When there are a finite number of discrete actions, this poses no problem. However, when the action space is continuous, this is highly non-trivial (and would be a very computational expensive subroutine).

\begin{algorithm}[!ht]
\caption{Twin Delayed DDPG (TD3)}
\label{td3_code}
\begin{algorithmic}[1]
\STATE Input: initial policy parameters $w$, Q-function parameters $\theta_{1}$, $\theta_{2}$, empty replay buffer $\mathcal{D}$
\STATE Set target parameters equal to main parameters \\ $w^{-}\leftarrow w$, $\theta_{1}^{-}\leftarrow\theta_{1}$, $\theta_{2}^{-}\leftarrow\theta_{2}$
\REPEAT
\STATE Observe state $x$ and select action \\ $u=\text{clip}(\hat{\pi}(x;w)+\epsilon,x_{\text{Low}},x_{\text{High}})$, where $\epsilon\sim\mathcal{N}$
\STATE Execute $u$ in the environment
\STATE Observe next state $x'$, reward $r$, and done signal $d$ to indicate whether $x'$ is terminal
\STATE Store $(x,u,r,x',d)$ in replay buffer $\mathcal{D}$
\STATE If $x'$ is terminal, reset environment state
\IF{it is time to update}
\FOR{$j$ in range(however many updates)}
\STATE Randomly sample a batch of transitions \\ $B=\{(x,u,r,x',d)\}$ from $\mathcal{D}$
\STATE Compute target actions
\begin{multline*}
    u'(x') = \text{clip}(\hat{\pi}(x';w^{-})  +\text{clip}(\epsilon,-c,c),\\x_{\text{Low}},x_{\text{High}}),\quad
    \epsilon\sim\mathcal{N}(0,\sigma)
\end{multline*}
\STATE Compute targets
\begin{equation*}
    q(r,x',d)=r+\gamma(1-d)\min_{i=1,2}\hat{Q}(x',u'(x');\theta_{i}^{-})
\end{equation*}
\STATE Update Q-functions by one step of gradient descent
\begin{equation*}
    \nabla_{\theta_{i}}\frac{1}{|B|}\sum_{(x,u,r,x',d)\in B}(\hat{Q}(x,u;\theta_{i})-q(r,x',d))^{2},
\end{equation*}
for $i=1,2$
\IF{$j$ mod $policy\,delay=0$}
\STATE Update policy by one step of gradient ascent
\begin{equation*}
    \nabla_{w}\frac{1}{|B|}\sum_{x\in B}\hat{Q}(x,\hat{\pi}(x;w);\theta_{1})
\end{equation*}
\STATE Update target networks
\begin{align*}
    \theta_{i}^{-} & \leftarrow (1-\tau )\theta_{i}^{-}+\tau \theta_{i},\quad\text{for }i=1,2 \\
    w^{-} & \leftarrow (1-\tau)w^{-}+\tau w
\end{align*}
\ENDIF
\ENDFOR
\ENDIF
\UNTIL{convergence}
\end{algorithmic}
\end{algorithm}

The deep deterministic policy gradient (DDPG) \cite{lillicrap2019} algorithm is specially adapted for environments with continuous action spaces. It uses neural networks to approximate both the Q-function and a deterministic policy, i.e. the policy network deterministically maps a state to a specific action. For exploration, one adds noise sampled from a stochastic process $\mathcal{N}$ to the actions of the deterministic policy and updates it by a gradient-based learning rule. As in deep Q-learning, the DDPG algorithm uses a replay buffer and target networks to improve stability during neural network training. Further details of the DDPG algorithm and its performance on different simulated physics tasks are given by Lillicrap et al. \cite{lillicrap2019}.

Although the DDPG algorithm is quite powerful, it has a direct successor, the Twin Delayed DDPG (TD3) \cite{fujimoto2018} algorithm, which further improves the stability by employing three critical tricks. The first trick addresses a particular failure mode of the DDPG algorithm: if the Q-function approximator develops an incorrect sharp peak for some actions, the policy will quickly exploit that peak and then have brittle or incorrect behavior. This failure mode can be averted by smoothing out the Q-function over similar actions. For this, one computes the action that is used to form the Q-learning target in the following way:
\begin{equation}
u'(x')=\text{clip}(\hat{\pi}(x';w^{-})+\text{clip}(\epsilon,-c,c),x_{\text{Low}},x_{\text{High}}),
\end{equation}
where $\epsilon\sim\mathcal{N}(0,\sigma)$ is noise sampled from a Gaussian process (target policy noise). The action is based on the target policy, but with clipped noise added (target noise clip $c$). After adding the noise, the target action is also clipped to lie in the valid action range ($x_{\text{Low}}, x_{\text{High}}$). The second trick is to learn two Q-functions $\hat{Q}(x,u;\theta_{i})$, for $i=1,2$, instead of one and use the smaller of the two Q-values to form the target. This improvement reduces overestimation in the Q-function.
\begin{equation}
q(r,x',d)=r+\gamma(1-d)\min_{i=1,2}\hat{Q}(x',u'(x');\theta_{i}^{-})
\end{equation}
The third trick is to update the policy less frequently than the Q-functions (policy delay) to damp the volatility that arises in the DDPG algorithm. Algorithm \ref{td3_code} shows the full pseudocode of the TD3 algorithm. The done signal $d$ is equal to one when $x'$ is the terminal state and otherwise equal to zero. The done signal guarantees that the agent gets no additional rewards after the current state at the end of an episode.

In addition to enhancements that improve the stability of the training process, research is also carried out to speed up the learning process of RL agents \cite{wang2020}. Besides DDPG, TD3, or SAC \cite{haarnoja2017}, which are so-called off-policy algorithms, there are also state-of-the-art on-policy algorithms like TRPO \cite{schulman2017a} or PPO \cite{schulman2017}. Nevertheless, on-policy methods are much more sample inefficient and have longer training time to achieve equivalent performances. From a control perspective, reinforcement learning converts the system identification problem and the optimal control problem to machine learning problems. Similar to explicit model predictive control it also addresses the problem of removing one of the main drawbacks of model predictive control, namely the need to solve a complex optimization problem online to compute the control action.\\

\noindent The main advantages of RL for control:
\begin{itemize}
    \item no derivation of a suitable state-space model, model order reduction or linearization needed
    \item direct use of a nonlinear simulation model
    \item ideal for highly dynamic situations (no complex online optimization needed)
    \item complex reward functions enable complicated goals
\end{itemize} \vspace{1mm}
The main disadvantage of RL for control:
\begin{itemize}
    \item stability of the controller is in general not guaranteed
\end{itemize}\vspace{1mm}
Concerning the last point (stability), we would like to make a remark. The controller's output can always be tested using the simulation environment, and there has been promising recent work on certifying stability of RL policies \cite{jin2018}.

\section{Simulation Environment and RL Implementation}
\label{sec:sim_environment}
A suitable simulation environment for our intended use is given by EcosimPro \cite{vila2018}. 
EcosimPro is a modeling and simulation tool for 0D or 1D multidisciplinary continuous and discrete systems. The system description is based on diﬀerential-algebraic equations and discrete events. Within a graphical user interface, one can combine diﬀerent components, which are arranged in several libraries. Of particular interest are the European Space Propulsion System Simulation (ESPSS) libraries, which are commissioned by the European Space Agency (ESA). These EcosimPro libraries are suited for the simulation of liquid rocket engines and have continuously been upgraded in recent years.

We use the TD3 implementation of Stable-Baselines \cite{stable-baselines}. Stable-Baselines is a set of improved implementations of RL algorithms based on OpenAI Baselines. It features a common interface for many modern RL algorithms and additional wrappers for preprocessing, monitoring, and multiprocessing. We encapsulate our simulation environment into a custom OpenAI Gym environment using an interface between EcosimPro and Python. Hence, we can directly use Stable-Baselines for training and testing. A big advantage of the RL approach is that it works regardless of whether one uses a lumped parameter model, continuous state-space models, surrogate models employing artificial neural networks \cite{waxenegger-wilfing2020}, \cite{dresia2019}, or a combination of the above.

\section{Test Case}
\label{sec:test_case}

\begin{figure}[!t]
\centerline{\includegraphics[width=\columnwidth]{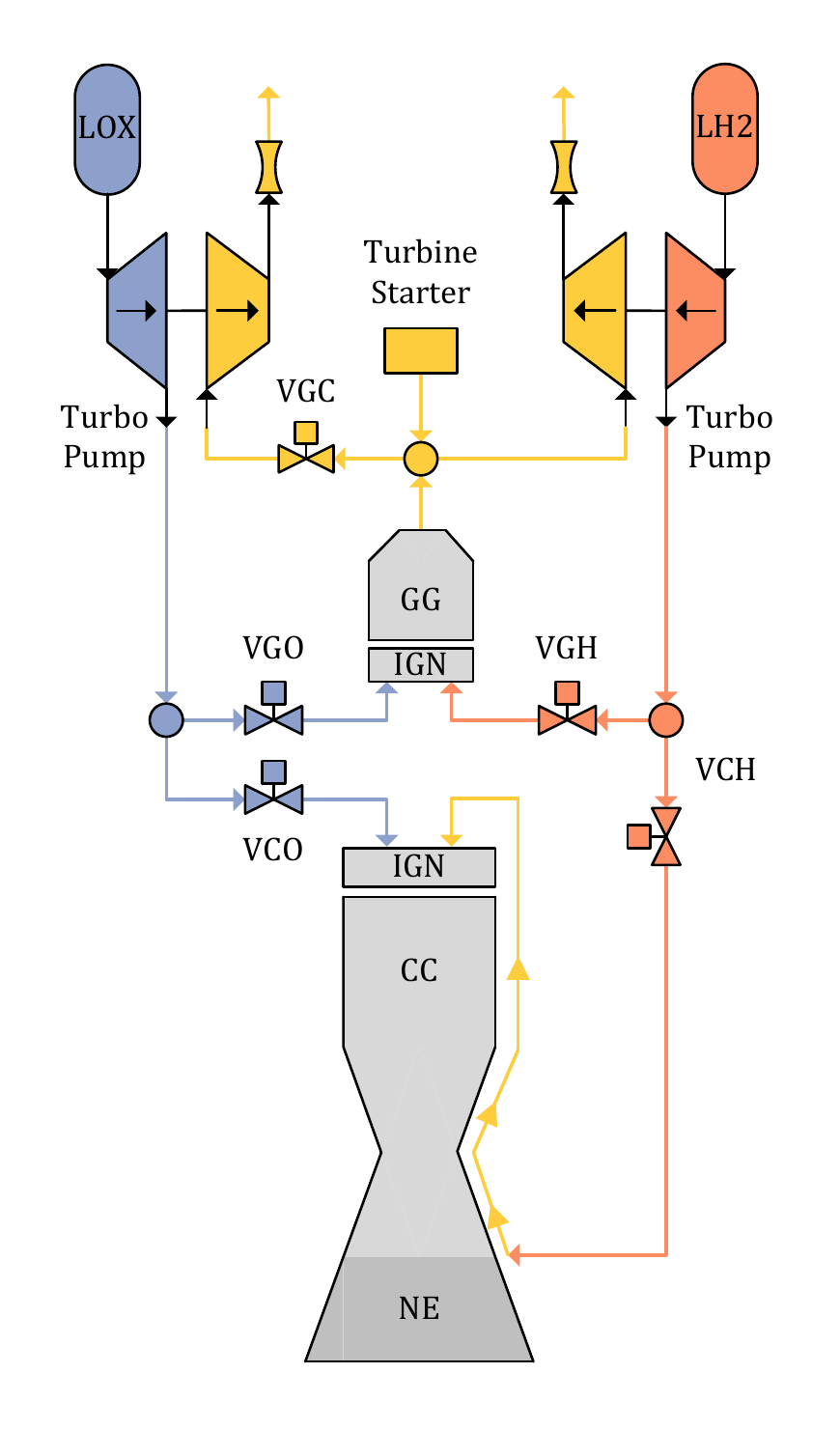}}
\caption{Flow plan of the considered engine architecture. Some of the propellants are burned in an additional combustion chamber, the gas-generator (GG), and the resulting hot-gas is used as the working medium of the turbines which power the engine's pumps. The gas is then exhausted. The engine architecture features five valves, but only three valves (VGH, VGO, VGC) are used for closed-loop control.}
\label{fig:scheme}
\end{figure}

\begin{figure*}[!t]
\centerline{\includegraphics[width=2\columnwidth]{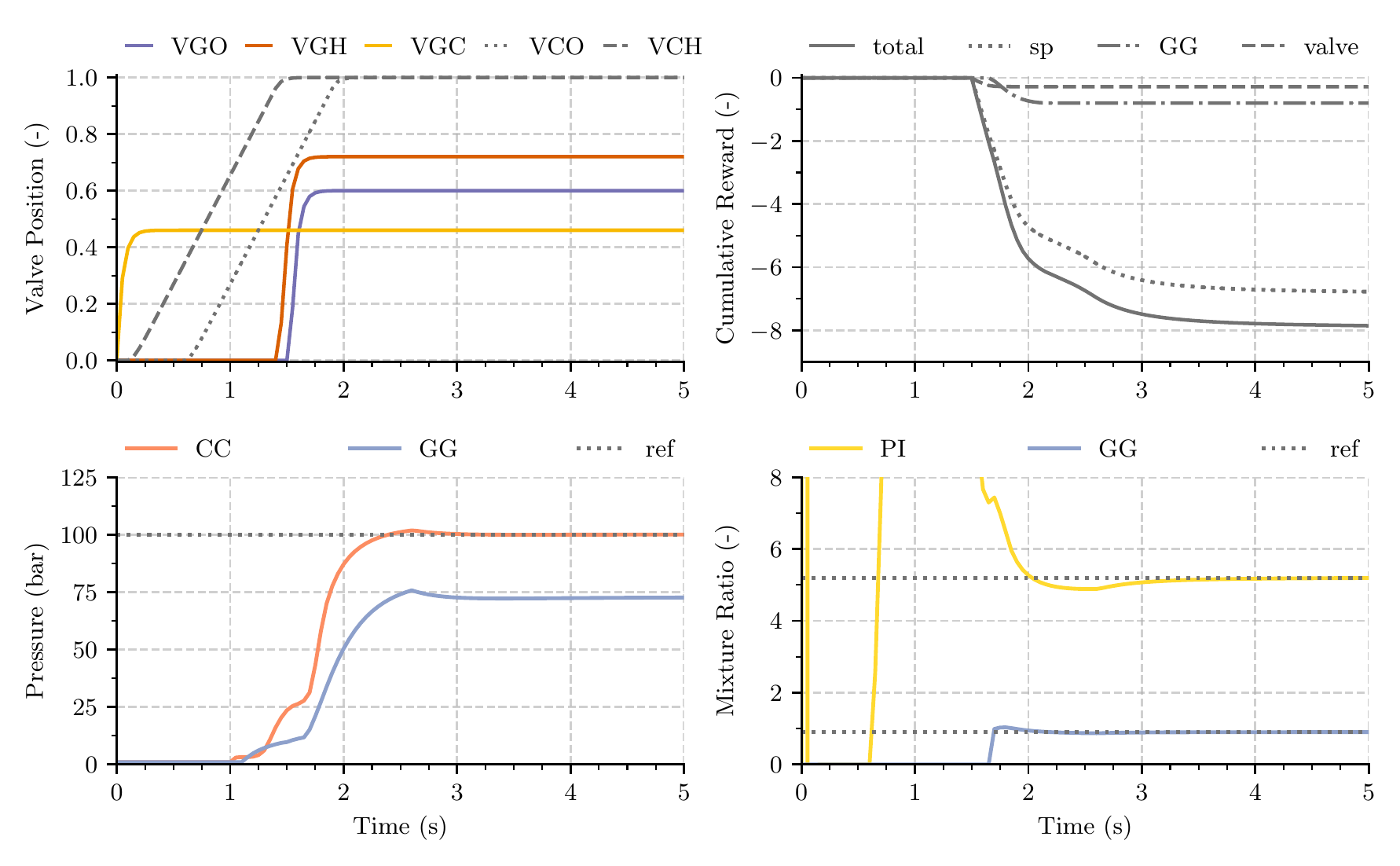}}
\caption{\SI{100}{\bar} nominal open-loop start-up sequence. The main combustion chamber pressure settles at \SI{100}{\bar}, while the gas-generator pressure reaches \SI{75}{\bar}. The reference mixture ratios are given by \num{5.2} and \num{0.9}. The engine reaches steady-state conditions after approximately \SI{4}{\second}.}
\label{fig:sequence_100bar}
\end{figure*}

The engine architecture considered to study the suitability of an RL approach for the control of the transient start-up is shown in Fig. \ref{fig:scheme}.  It is similar to the architecture of the European Vulcain 1 engine \cite{iffly1999}, which powered the cryogenic core stage of Ariane 5 launch vehicle before it got replaced by the upgraded Vulcain 2 engine. It is fed with cryogenic liquid oxygen (LOX) at a temperature of \SI{92}{\kelvin} and liquid hydrogen (LH2) at \SI{22}{\kelvin}. The engine generates approximately \SI{1}{\mega \newton} of thrust at a main combustion chamber (CC) pressure of \SI{100}{\bar} and a chamber mixture ratio of \num{5.6}, i.e. the chamber mass flow of the oxidizer divided by the chamber mass flow of the fuel equals \num{5.6}. The engine cycle is an open gas-generator cycle, where a small amount of the propellants is burned in a small combustion chamber, the gas-generator (GG). The gas-generator is operated at a fuel-rich mixture ratio of \num{0.9}. The produced hot-gas is used to drive the turbines before it is exhausted. The turbines power the pumps which force the propellants into the combustion chambers. LH2 is used to cool the nozzle and main combustion chamber before it gets burned. A convergent-divergent nozzle, which usually includes an uncooled nozzle extension (NE), accelerates the combustion gases to generate thrust.

The actuators are given by five flow control valves (VCO, VCH, VGO, VGH, VGC). VCO and VCH are the main combustion chamber valves that regulate the propellant flow to the combustion chamber. VGO and VGH, the gas-generator valves, are used to control the gas-generator pressure and mixture ratio. The turbine valve, VGC, is located downstream of the gas-generator and is used to determine the hot-gas flow ratio between the LOX and LH2 turbines. Thus, this valve mainly influences the global mixture ratio (PI, pump-inlet). Further actuators are the ignition systems (IGN) for the main combustion chamber and the gas-generator, as well as a turbine starter. The turbine starter produces hot-gas for a short period to spin up the turbines during the start-up.

To start the engine and reach steady-state conditions, a succession of discrete events, including valve openings and chamber ignitions are necessary. The start-up sequence of an engine, i.e. the chronological order of oxidizer and fuel valve openings, as well as the precise ignition timings, determines the engine's thermodynamic conditions and mechanical stresses during start-up. A non-ideal start-up sequence can damage the engine, e.g. by excessive temperatures. These high temperatures can substantially damage the turbine blades or at least reduce their live expectancy \cite{ryan1986}. An optimal start-up sequence leads to a smooth ignition of the combustion chamber and gas-generator with low thermal and mechanical stresses. An open-loop start-up sequence (OLS) for a steady-state chamber pressure of \SI{100}{\bar} is shown in Fig. \ref{fig:sequence_100bar}. The sequence does not correspond exactly to the Vulcain 1 start-up sequence, but it is realistic for such an engine cycle. The flow control valves are opened monotonically until the end positions are reached. First the VCH valve starts to open at $t=\SI{0.1}{\second}$, followed by VCO at $t=\SI{0.6}{\second}$. A fuel-lead transient is usually used for a smooth ignition of the combustion chamber, which takes place at $t=\SI{1.0}{\second}$. At this point, the main combustion chamber is burning at low pressure, only fed by the tank pressurization. At $t=\SI{1.1}{\second}$, the turbine starter activates to spin up the turbopumps, which start to build up the pressure in the main combustion chamber and at the gas-generator valves VGO, VGH. At $t=\SI{1.4}{\second}$ and $t=\SI{1.5}{\second}$, the gas-generator valves VGH and VGO open and the gas-generator is ignited. The VGC valve is set to a fixed position during the entire start-up sequence. At $t=\SI{2.6}{\second}$, the turbine starter is burned out and the engine reaches steady-state conditions after approximately \SI{4}{\second}. The valve positions in Fig. \ref{fig:sequence_100bar} are tuned to reach a main combustion chamber pressure $p_{\text{cc}}$ of \SI{100}{\bar}, a global mixture ratio $\MR_{\mathrm{PI}}$ of \num{5.2} and a gas-generator mixture ratio $\MR_{\mathrm{GG}}$ of \num{0.9}.

Although RL can solve discrete or hybrid control problems, there are controllability and observability issues during the first phase of discrete events due to very low mass flows \cite{nemeth1991}. Thus we focus on the fully continuous phase starting at $t=1.5s$. The goal of the controller (agent) is to drive the engine as fast as possible towards the desired reference by adjusting the flow control valve positions. In our multi-input multi-output (MIMO) control tasks, only three flow control valves, VGO, VGH, and VGC, are used for active control of the combustion chamber pressure, and the mixture ratio of the gas-generator as well as the global mixture ratio. The valve actuators are modeled as a first-order transfer function with a time constant of $\tau = \SI{0.05}{\second}$ and a linear valve characteristic. The minimum valve position is set to \num{0.25} for VGH and  VGO and \num{0.20} for VGC, respectively. The maximum valve position is \num{1.0} for all valves.

We study different reference values for the combustion chamber pressure, namely \num{80} and \SI{100}{\bar}. The reference mixture ratios remain the same, \num{5.2} for the global mixture ratio and \num{0.9} for the mixture ratio of the gas-generator. For a combustion chamber pressure of \SI{80}{\bar}, the valve timings are the same, but the final valve positions were adjusted accordingly (see Fig. \ref{start_up80}). Furthermore, we study the effect of degrading turbine efficiencies on the start-up transient. This scenario has practical relevance for future reusable engines. The use of cryogenic propellants leads to significant thermostructural challenges in the operation of turbopumps. Since thermal stresses depend on the temperature gradient, they can cause significant loads on the metal parts that have to react to these stresses. The resulting fatigue deformation \cite{ryan1986} affects the performance of the turbines. Furthermore, the aging of seals can cause increased leakage mass flows, which in turn decreases the turbine efficiency \cite{roelke1973}. Additional reasons are turbine blade erosions \cite{hampson1984} and soot depositions on the turbine nozzles by fuel-rich gases when using hydrocarbons as fuel. These soot depositions can decrease the effective nozzle area up to \SI{20}{\percent} \cite{meland1989}, thus reducing the turbopump performance. Furthermore, soot depositions are a main shortcoming for reusable engines due to the unpredictable impacts for engine re-start \cite{pempie2001}. To study the effect of degrading turbine efficiencies for our generic test case, we simulate and evaluate the performance of the open-loop start-up sequence, a family of PID controllers, and our RL-agent for 16 different combinations of LOX and LH2 turbine efficiencies. For each turbine, 4 different efficiencies are considered ranging from \SI{100}{\percent} to \SI{85}{\percent} of the nominal value.

The reward, which is used to evaluate a start-up sequence and to train the RL agent consists of 3 different terms:
\begin{equation}
    r = r_{\mathrm{sp}} + r_{\mathrm{GG}} + r_{\mathrm{valve}}.
\end{equation}
The first term
\begin{equation}
    r_{\mathrm{sp}} = -\sum_{x_\mathrm{i}}^{} \mathrm{clip} \left( \left|\frac{x_\mathrm{i} - x_{i,ref}}{x_{i,ref}}\right|, 0.2 \right)
\end{equation}
for $x_\mathrm{i\,} \in [p_{\mathrm{CC}},\,{\MR}_{\mathrm{GG}},\,{\MR}_{\mathrm{PI}}]$ penalizes deviations from the desired set-point for all controlled variables. Each reward component in this term is clipped to a maximum value of \num{0.2} to improve training and to balance the accumulated reward during start-up and steady-state. The second term of the reward
\begin{equation}
    r_{\mathrm{GG}} = -
    \begin{dcases}
    \frac{{\MR}_{\mathrm{GG}} - {\MR}_{\mathrm{GG,ref}}}{{\MR}_{\mathrm{GG,ref}}}, & \text{if }  \frac{{\MR}_{\mathrm{GG}}}{{\MR}_{\mathrm{GG, ref}}} > 1 \\
     0, & \text{otherwise}
    \end{dcases}
\end{equation}
additionally penalizes high mixture ratios in the gas-generator. High mixture ratios are dangerous because they result in increased temperatures and thus possible damaging conditions to the turbines. The last reward component
\begin{equation}
    r_{\mathrm{valve}} = -\frac{\left|s_\mathrm{VGH}\right| + \left|s_\mathrm{VGO}\right| + \left|s_\mathrm{VGC}\right|}{3},
\end{equation}
where $s$ is the change in valve position between two time steps, penalizes excessive valve motion. By adding this component, we encourage the agent to move the valves as little as possible to avoid valve wear, valve oscillations, and valve jittering. All together, this reward allows the agent to trade off between reaching the desired reference point as fast as possible, avoiding steady-state errors, minimizing overshoots, and reducing valve motion as much as possible. Fig. \ref{fig:sequence_100bar} shows all 3 components of the cumulative reward for the nominal OLS for \SI{100}{\bar}. Since the valves are only moved once in the OLS, the contribution of $r_{\mathrm{valve}}$ to the total reward is low. As the overshoot in the gas-generator mixture ratio is also small (small $r_{\mathrm{GG}}$), the total reward is mainly composed of the set point error $r_{\mathrm{sp}}$.

To train and use a RL agent, one needs to define the observation and action space of the agent. The observation space, i.e. the variables the agent receives from the environment at each time step, should at least contain sufficient information to unambiguously define the state of the system. In our set-up, the observation space
\begin{multline}
    X = [p_\mathrm{cc,ref},\,\epsilon_\mathrm{cc},\,\epsilon_\mathrm{PI},\,\epsilon_\mathrm{GG},\, Pos_\mathrm{VGO},\,Pos_\mathrm{VGH},\,Pos_\mathrm{VGC},\\\omega_\mathrm{LOX},\,\omega_\mathrm{LH2}]
\end{multline}
 contains 9 variables, where $\epsilon_\mathrm{i} = x_\mathrm{i} - x_\mathrm{i,ref}$ is the absolute error for each controlled variable, $Pos_\mathrm{VGO}, Pos_\mathrm{VGH}, \text{ and } Pos_\mathrm{VGC}$ are the positions of all control valves, and $\omega_{\mathrm{LOX}} \text{ and } \omega_{LH2}$ are the rotational speeds of the turbopumps. The observation space is normalized with the reference steady-state values. All variables in our observation state are measurable in real engines. Thus our approach is not limited to simulation environments, where one could possibly use variables that are impossible to measure directly in real engines (e.g. the turbine efficiencies). The agent's action space $U$ consists of all 3 gas-generator valve positions
 \begin{equation}
    U = [Pos_\mathrm{VGO},\,Pos_\mathrm{VGH},\,Pos_\mathrm{VGC}].
\end{equation}
At each time step, the RL agent receives observations from the environment and sends control signals to the flow control valves of the engine. The frequency of interaction between the controller (RL-agent and PID) and the environment is set to \SI{25}{\hertz}.\\


\section{Results}
\label{sec:results}
In this section, we assess the performance of our RL controller. For this we use the approximation of the integrated absolute error over one entire episode for each controlled variable:

\begin{equation}
    (\IAE)_{i} = \int_{}^{}\left|\epsilon_i\right| dt \approx \sum_{t_j}^{}\left|\epsilon_i(t_j)\right|,
\end{equation}
where $t_j$ are the discrete time steps. Furthermore, we evaluate the average steady-state values of the controlled variables from $t=\SI{3.5}{\second}$ to $t=\SI{5.0}{\second}$ and the value of the cumulative reward.

\begin{figure}
\centerline{\includegraphics[width=\columnwidth]{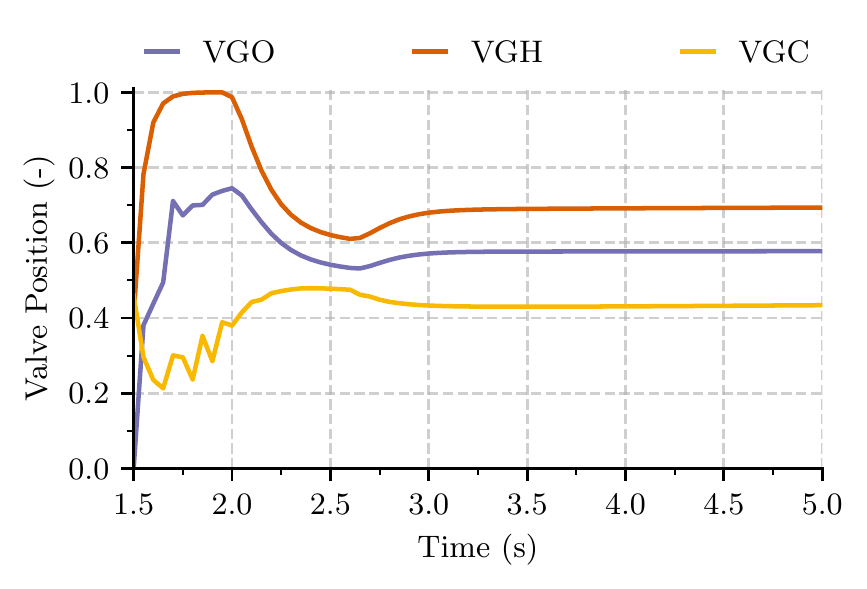}}
\caption{Manipulated valve positions by the PID controllers for the \SI{100}{\bar} nominal start-up. VGO is used to control the mixture ratio of the gas-generator, while VHG and VGC control the pressure of the main combustion chamber and the global mixture ratio respectively.}
\label{fig:pid_valve_positions}
\end{figure}

\begin{figure}
\centerline{\includegraphics[width=\columnwidth]{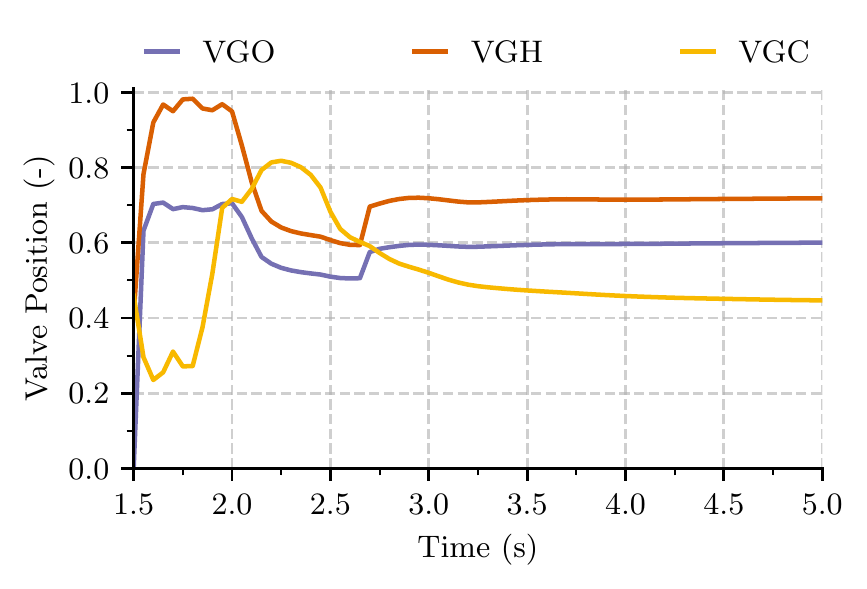}}
\caption{Manipulated valve positions by the RL agent for the \SI{100}{\bar} nominal start-up. The action clearly changes at $t=\SI{2.6}{\second}$, which is the time when the firing of turbine starter stops.}
\label{fig:rl_valve_positions}
\end{figure}

\begin{figure*}[p]
\centerline{\includegraphics[width=2\columnwidth]{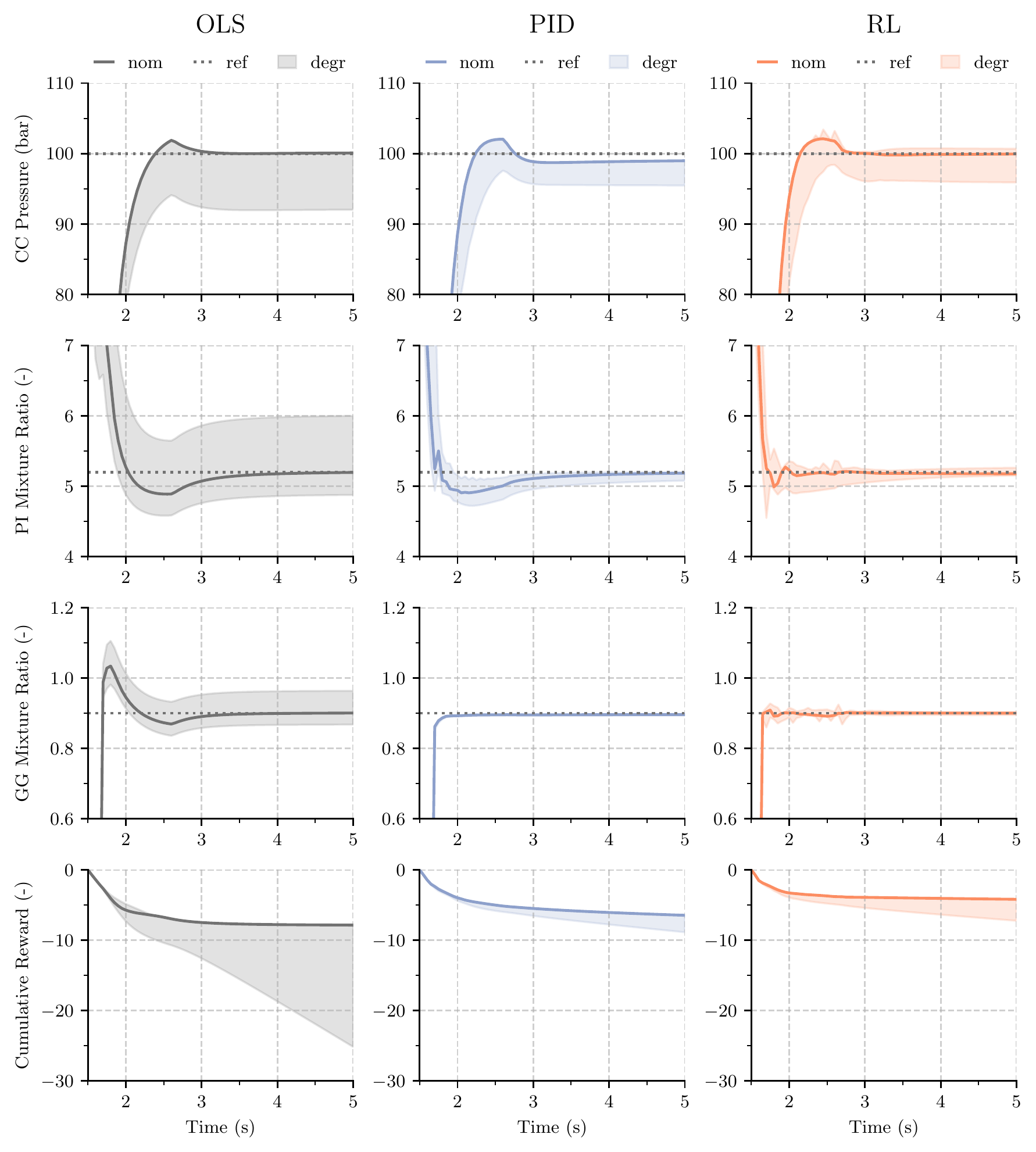}}
\caption{Comparison of the controlled variables for the 100 bar start-up. Shaded area marks the range of the controlled variable for different degraded efficiencies. At different turbine efficiencies the standard open-loop sequence provides significantly different steady-state values for the chamber pressure and the mixture ratios.}
\label{fig:comparison_100bar}
\end{figure*}

Before we turn to the performance of closed-loop control, let us record the downsides of open-loop sequences (OLS). The first column in Fig. \ref{fig:comparison_100bar} shows the resulting engine start-up for the nominal OLS and degrading turbine efficiencies. For the latter, the steady-state values deviate strongly from the reference values. The minimum steady-state value of the main combustion chamber pressure is \SI{92}{\bar}. The steady-state of the global mixture ratio varies between \num{4.9} and \num{6.0}. To prevent fuel or oxidizer from running out during a mission in the event of a persisting mixing ratio deviation, the loaded propellants must be increased, which reduces the payload capacity of the launch vehicle. A further negative effect is that the temperature in the combustion chamber can rise significantly due to a shift in the mixing ratio, which could reduce the engine's service life. Additionally, the steady-state value of the mixture ratio of the gas-generator changes too. The temperature in the gas-generator is sensitive to the mixture ratio, and an increased temperature can also damage the turbines. These damaging conditions are especially problematic for reusable engines, which must possess a long service life. The same implications apply to the \SI{80}{\bar} case as Fig. \ref{fig:comparison_80bar} shows.

Those unfavorable effects can be counteracted with a closed-loop control system. First, we tune a family of PID controllers to achieve the start-up. The process of controlling the chamber pressure of the main combustion chamber, the mixture ratio of the gas-generator, and the global mixture ratio by manipulating VGO, VGH, and VGC is coupled. E.g. changing VGO does affect not only the mixture ratio of the gas-generator but also the other two controlled variables. Nevertheless, for rocket engine control near steady-state conditions, the standard approach is to use separate PID controllers and tune the control loops at different speeds to avoid oscillations \cite{lorenzo1992}. Hence, we also use three separate controllers. 

The first controller manipulates VGO to control the mixture ratio of the gas-generator, the second controller manipulates VGH to control the chamber pressure of the main combustion chamber, and the third controller manipulates VGC to control the global mixture ratio. Starting far away from the reference point can be problematic for a simple PID controller because the integrator begins to accumulate a significant error during the rise. Consequently, a large overshoot may occur. Modern PID controllers use different methods to address this problem of integrator-windup. We use a simple feedback loop, where the difference between the actual and the commanded actuator position is fed back to the integrator, to avoid the effects of saturation. If there is no saturation, our anti-windup scheme has no effect. The ratio between the time constant for the anti-windup and the integration time is \num{0.1} for all PID controllers. 

For PID parameter tuning, we directly use the simulation model coupled with a genetic algorithm \cite{cominos2002} of the Distributed Evolutionary Algorithms in Python (DEAP) framework \cite{DEAP_JMLR2012}. To guarantee a fair comparison, we use the reward function to calculate the fitness value of a certain parameter combination. Table \ref{pid} presents the optimal PID parameters, which maximize the reward function. The genetic algorithm uses a population of \num{5000} valid individuals and evolves the population for \num{20} generations. Fig. \ref{fig:pid_valve_positions} show that the best PID controllers open the valves in a nonmonotonic way, which leads to a faster start-up. Furthermore, the PID controllers fulfill their main task: the feedback loops lead to an adjustment of the valve positions at lower turbine efficiencies and significantly reduce the deviations from the reference values of the controlled variables. Due to the structure of PID controllers, with their proportional, integral, and derivative terms, the shape of the control input is restricted and does not provide optimal control. 

Fig. \ref{fig:comparison_100bar} shows that the optimized PID controllers lead to certain overshoots of the main combustion chamber pressure and the global mixture ratio. It is possible to eliminate the overshoots by changing the PID parameters, but this would significantly increase the settling time. For our parameters, there is still an error in combustion chamber pressure after \SI{4}{\second} even for nominal efficiencies. The settling time is not the only reason for a large error in the combustion chamber pressure in the case of lower turbine efficiencies. For the lowest turbine efficiencies, a combustion chamber pressure of \SI{100}{\bar} is physically no longer possible while maintaining the other constraints (especially the desired gas-generator mixture ratio). A specific disadvantage of PID controllers is that degenerating efficiencies or other system parameters cannot be considered directly as further input variables. Fig. \ref{fig:pid_valve_positions_80} shows that for the \SI{80}{\bar} start-up VGC oscillates a little. It is challenging to tune a single family of PID controllers for different reference combustion chamber pressures. For even lower combustion chamber pressures (deep throttling), it becomes more and more difficult to achieve a convincing performance for all operating conditions. The prevention of oscillations leads to an increased settling time for all reference values. All in all, the performance of the PID controllers is not perfect but satisfactory for the case of \num{100} and \SI{80}{\bar} and fixed mixture ratios.

\begin{table}
\centering
\renewcommand{\arraystretch}{1.3}
\caption{Controller Performance for Nominal Turbine Efficiencies}
\label{table1}
\setlength{\tabcolsep}{2.3pt}
\begin{tabular}{r c c c r c rrr c rrr}
\toprule
\toprule
   Target      & &  \mc{Algo.}  & & \mc{Reward} & & \multicolumn{3}{c}{Steady-State Values} & & \multicolumn{3}{c}{\IAE} \\
 \cmidrule{1-1}
 \cmidrule{3-3}
 \cmidrule{5-5}
 \cmidrule{7-9}
  \cmidrule{11-13}
$p_{\mathrm{CC}}$ & & & & \mc{cum.} & & $p_{\mathrm{CC}}$ & $\MR_{\mathrm{GG}}$ & $\MR_{\mathrm{PI}}$ && $p_{\mathrm{CC}}$ & $\MR_{\mathrm{GG}}$  & $\MR_{\mathrm{PI}}$ \\
\mc{(\si{\bar})}   &  &  & & \mc{(--)} & & \mc{(\si{\bar})} & \mc{(--)} & \mc{(--)} && \mc{(\si{\bar})} & \mc{(--)} & \mc{(--)}\\
\midrule
100  & & OLS  & & -7.9  & & 100.0  & 0.90  & 5.18 & & 591   & 4.7   & 27 \\ 
    & & PID    && -6.5 & & 98.9 & 0.90  & 5.17 & & 632   & 4.0   & 19  \\ \vspace{1mm}
    & & RL    &&  -4.2   &  & 99.9 &0.90       &    5.18   &  & 519&2.8    & 13  \\ 
80 & & OLS   &  & -7.0  & & 79.7  & 0.90 & 5.20  & & 576  &  6.0  & 15  \\ 
    && PID   &&   -5.2    & & 79.9     &     0.90  &     5.20  & & 433     &    3.8   & 9\\ 
    && RL        && -4.4      & & 80.8     &    0.90   &     5.18  & &  366    &     2.9  & 8 \\
\bottomrule
\end{tabular}
\end{table}

\begin{table*}
\centering
\renewcommand{\arraystretch}{1.3}
\caption{Controller Performance for 16 Different Combinations of Degraded Turbine Efficiencies}
\label{table2}
\setlength{\tabcolsep}{3pt}
\begin{tabular}{rc c c rr c  rrrr  rrrr  rrrr c  rr  rr  rr}
\toprule
\toprule
   \mc{Target} &     &  Algo. && \multicolumn{2}{c}{Reward} &  & \multicolumn{12}{c}{Steady-State Values} && \multicolumn{6}{c}{Integral Absolute Error \IAE}  \\
 \cmidrule{1-1}
  \cmidrule{3-3}
\cmidrule{5-6}
 \cmidrule{8-19}
  \cmidrule{21-26}
 \mc{$p_{\mathrm{CC}}$ (\si{\bar})} & && & \multicolumn{2}{c}{cumulative (--)} & & \multicolumn{4}{c}{$p_{\mathrm{CC}}$ (\si{\bar})} & \multicolumn{4}{c}{$\MR_{\mathrm{GG}}$ (--)} & \multicolumn{4}{c}{$\MR_{\mathrm{PI}}$ (--)}&&  \multicolumn{2}{c}{$p_{CC}$ (\si{\bar})} & \multicolumn{2}{c}{$\MR_{\mathrm{GG}}$ (--)} & \multicolumn{2}{c}{$\MR_{\mathrm{PI}}$ (--)}\\
 & &&& mean & sd && min & max & mean & sd & min & max & mean & sd &min & max & mean & sd &&  mean & std & mean & std & mean & std \\
\midrule
100  && OLS  && -14.9 & 5.2 && 92.0  & 100.0  & 96.1 & 2.3 & 0.87  & 0.96  & 0.91 & 0.03  & 4.9   & 6.0   & 5.4 &0.3 && 1841 & 151 & 5.8 & 1.0 & 45  & 18\\ 
  && PID  && -7.6 & 0.7 && 95.5 & 98.9 & 97.7 & 1.0 & 0.89 & 0.90 & 0.89 & 0.01 & 5.0 & 5.2 & 5.2 &  0.1 && 758& 98& 4.1& 0.1&25&4  \\ \vspace{1mm}
  && RL  && -5.5 & 0.9 && 96.0&100.7&98.8&1.4&0.89&0.90&0.90&0.02&5.1&5.3&5.2&0.0&&636&91&2.9&0.1&29&6\\ 
80  && OLS && -15.4&5.8&& 71.1 & 79.7 & 75.5 & 2.4 & 0.86 & 0.96 & 0.91 & 0.03& 4.8 & 6.2 & 5.5 & 0.4&& 815 & 149 & 7.0& 0.9& 36& 21\\ 
  && PID  && -5.7&0.6&& 79.5& 79.9& 79.7&0.1& 0.90 & 0.90 & 0.90 &0.00& 5.2 & 5.5 & 5.2 &0.1&& 459 & 18 & 3.9 & 0.1 & 12 & 6 \\ 
  && RL   &&  -4.4&0.8&& 79.6 & 82.1 & 80.3 & 0.6 & 0.89& 0.90 & 0.90 & 0.00 & 5.2 & 5.4 & 5.2 & 0.0&& 366&30&3.3 &0.1& 10 & 5 \\ 
\bottomrule
\multicolumn{26}{c}{For each turbine, 4 different efficiencies are considered ranging from \SI{100}{\percent} to \SI{85}{\percent} of the nominal value.}
\end{tabular}
\end{table*}

Now we examine the performance of our RL approach. The comparison of Fig. \ref{fig:pid_valve_positions} and Fig. \ref{fig:rl_valve_positions} shows that at first glance the RL agent's behavior shows strong similarities to the PID controllers. The flow control valves are opened in a nonmonotonic way. Nevertheless, the agent can guarantee an even faster start-up, as presented in Fig. \ref{fig:comparison_100bar}. The RL controller can better control the combustion chamber pressure and the global mixing ratio. The control of the gas-generator mixture ratio is comparatively good. Furthermore, the RL agent can directly take the firing of the turbine starter into account. The action changes at $t=\SI{2.6}{\second}$, which is the time when the firing of turbine starter stops. Similar to the PID controllers, the RL agent can handle degrading turbine efficiencies to a certain extent. It can detect deviating efficiencies because the relationship between valve positions and controlled variables changes, and adjusts the start-up. A prerequisite for this is that the valve positions are also included in the observation space, and that experiences with different efficiencies were generated during the training.

Table \ref{table1} compares the rewards, steady-state values, and \IAE s of the studied approaches for nominal turbine efficiencies and both main combustion chamber pressures of \SI{100}{\bar} and \SI{80}{\bar}. The open-loop sequences are satisfying for the nominal start-ups. Nevertheless, both \IAE s and rewards show that improvement is possible. One can start up faster if the valves are opened nonmonotonously. Why is this not done for realistic start-up sequences? As already mentioned, it is common practice to determine the control sequences employing tests on test benches, which is expensive and time-consuming. With non-reusable engines, the demands on the control system are not so dramatic, and one can accept good but not optimal sequences as long as a large amount of development costs is saved. Another reason is that, as a rule, disturbances influence the start-up anyway and cancel out the advantages of optimized sequences. The advantages can only be realized by closing the control loop. The tuned PID controllers are better than the open-loop sequences concerning the value of the reward. The RL agent is even better. The RL agent and the PID controllers also achieve decent steady-state values.

Table \ref{table2} compares the rewards, steady-state values, and \IAE s of the studied approaches for degrading turbine efficiencies. We present the mean and standard deviation of the measures instead of giving all values for the 16 different combinations of turbine efficiencies. For the steady-state values, the minimum and maximum values are also listed in Table \ref{table2}. As already seen in Figure \ref{fig:comparison_100bar}, the OLS results in large deviations for degrading turbine efficiencies. For \SI{100}{\bar}, the steady-state main combustion chamber pressure ranges between \SIrange{92}{100}{\bar}. Furthermore, degrading turbine efficiencies strongly influence the overall mixture ratio $\MR_{\text{PI}}$. Large deviations in $\MR_{\text{PI}}$ (here from \SIrange{4.9}{6.0}{}) poses two major problems. First, the fuel and oxidizer tank volumes are designed for the nominal mixture ratio. Deviations in $\MR_{\text{PI}}$ result in a non-ideal utilization of the propellants, thus lowering the launcher's performance. Second, the mixture ratio in the main combustion chamber is directly affected by the overall mixture ratio, potentially resulting in more damaging conditions for the main combustion chamber. The cumulative reward for the OLS increases to a mean value of \num{-14.9} with a large standard deviation of \num{5.2}.

The controller performances of both closed-loop controllers highlight the benefits of closed-loop control for degrading turbine efficiencies. The mean and standard deviations of the cumulative rewards are much smaller for the PID controllers and the RL agent. The additional reduction for the RL agent is mainly due to an even faster start-up. The mean steady-state value of $p_{\text{CC}}$ is given by \SI{96.0}{\bar} for the agent, which is a little bit closer to \SI{100.0}{\bar} than the value of PID controllers and much closer than the value of the open-loop sequence. Furthermore, the maximum deviation is the smallest. The advantages of closed-loop control and especially the RL approach are also reflected in the mixture ratios, which are much closer to their nominal values compared to the OLS. The \IAE s also show that the RL agent performs better than the PID controllers.

\section{Conclusion and Outlook}
\label{sec:conclusion}
In this work, we presented a RL approach for the optimal control of the fully continuous phase of the start-up of a gas-generator cycle liquid rocket engine. Using a suitable engine simulator, we employed the TD3 algorithm to learn an optimal policy. The policy achieves the best performance compared with carefully tuned open-loop sequences and PID controllers for different reference states and varying turbine efficiencies. Furthermore, the prediction of the control action takes only \SI{0.7}{\milli \second}, which allows a high interaction frequency, and in comparison to MPC enables the real-time use of RL algorithms for closed-loop control. The modest computational requirements should be met by the current generation of engine control units. A potential drawback of the RL approach is the lack of stability guarantees.  Nevertheless, the control system can be tested using a high fidelity simulation model, and there is ongoing work on certifying stability of RL policies \cite{jin2018}.

The present work can be improved in many directions. It is necessary to carefully examine the performance of the controller when various disturbances occur. Disturbance rejection, integration of filtering, and observer design will be the focus of future work. Furthermore, even the most sophisticated models usually have prediction errors due to not included effects or model miss-specifications. Therefore, it is essential to ensure that controllers trained in a simulation environment are robust enough to be used in real applications. There are RL approaches that explicitly consider modeling errors. Domain randomization \cite{tobin2017} can produce agents that generalize well to a wide range of environments.
Another issue with RL is implementing hard state constraints. Using the example of liquid rocket engine control, one would like to impose hard constraints to limit the maximum rotational speed of the turbopumps and maximum temperatures to prevent damage to the engine. It is possible to approximate hard state constraints by carefully tuning the reward function, e.g. one can give the agent a sizeable negative reward upon constraint violation and possibly terminate the training episode. Besides, there has been recent work on implementing hard constraints in RL using constrained policy optimization \cite{achiam2017}. 

We would like to conclude this publication with an outlook on the potential advantages of this approach for rocket engine control. Controllers trained with RL can depend on many input variables, can be used for very different operating conditions, and can include multiple objectives. The thrust control of rocket engines is crucial for improving the performance of the launch vehicle, but it is particularly critical when using rocket engines for the soft landing of returning rocket stages. Deep throttling domains of an engine, i.e. 25-100\% range of nominal thrust, are not supposed to pose a problem for RL controllers. Regarding multiple objectives, one can modify the reward function to optimize both the system's performance and damage mitigation \cite{ray1994}. The coupling of sophisticated health monitoring systems, possibly based on machine learning techniques, with suitable policies trained by RL, can increase the reliability of launch systems further. Given a suitable simulation environment, end-to-end RL may even enable the training of integrated flight and engine control systems. Overall, it is hoped that the current work will serve as a basis for future studies regarding the application of RL in the field of rocket engine control.

\section*{Acknowledgment}
The authors would like to thank Wolfgang Kitsche, Robson Dos Santos Hahn, and Michael B\"orner for valuable discussions concerning the start-up of a gas-generator cycle liquid rocket engine.

\clearpage

\appendices

\section{Implementation and Training Details}
\label{sec:details}
The agent is trained for \num{100000} time steps, which is equal to approximately \num{1.5} hours of simulation time. The agents' hyperparameters are tuned with Optuna \cite{optuna_2019} and are presented in Table \ref{td3}. For exploration we use action noise sampled from an Ornstein-Uhlenbeck process \cite{uhlenbeck1930}. Table \ref{pid} shows the parameters of all three PID controllers and the corresponding controlled variable and control valve.

\begin{table}[h]
\centering
\renewcommand{\arraystretch}{1.3}
\caption{TD3 Hyperparameters}
\label{td3}
\begin{tabular}{ll}
\toprule
\toprule
Parameter      & Value \\
\midrule
number of hidden units per layer  & [400, 300] \\ 
number of hidden layer  & 2 \\ 
activation function & $ReLU$ \\
optimizer & Adam \\
number of samples per minibatch & 256 \\
learning rate & 0.001 \\
soft update coefficient ($\tau$) & 0.005 \\
train frequency & 10 \\
gradient steps & 10 \\
discount rate ($\gamma$) & 0.90 \\
warm-up steps & \num{5000} \\
total training steps & \num{100000} \\
size of the replay buffer & \num{25000} \\
target policy noise & 0.01 \\
target noise clip & 0.02 \\
policy delay & 2 \\
action noise type & Ornstein-Uhlenbeck \\
action noise std ($\sigma$) & 0.05 \\
rate of mean reversion ($\theta$) & 0.25 \\
\bottomrule
\end{tabular}
\end{table}

\begin{table}[h]
\centering
\renewcommand{\arraystretch}{1.3}
\caption{PID Parameters}
\label{pid}
\begin{tabular}{cccl}
\toprule
\toprule
Valve & Controlled Variable & Parameter      & Value \\
\midrule
VGO  &   ${\MR}_\mathrm{GG}\,(-)$ & $K_p$  & 98.5 \\ 
    & &$T_i$  & 36.3 \\
    & &$T_d$ & \num{3.56e-4} \\
\midrule
VGH  & $p_\mathrm{cc}\,(\si{\pascal})$& $K_p$  & \num{2.59e-7} \\ 
    && $T_i$  & 1.22 \\
    & &$T_d$ & \num{6.82e-3} \\
\midrule
VGC  & ${\MR}_\mathrm{PI}\,(-)$ & $K_p$  & 0.786 \\ 
    & &$T_i$  & 1.06 \\
    & &$T_d$ & \num{2.12e-2} \\
\bottomrule
\end{tabular}
\end{table}

\vfill\eject
\section{Plots for 80 bar case}
\begin{figure}[h]
\centerline{\includegraphics[width=\columnwidth]{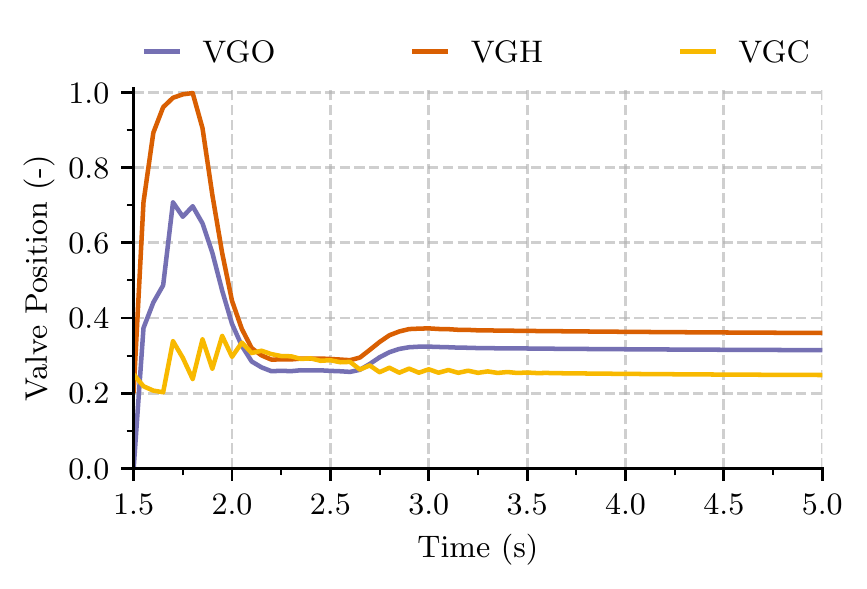}}
\caption{Manipulated valve positions by the PID controllers for the \SI{80}{\bar} nominal start-up. VGO is used to control the mixture ratio of the gas-generator, while VHG and VGC control the pressure of the main combustion chamber and the global mixture ratio respectively.}
\label{fig:pid_valve_positions_80}
\end{figure}

\begin{figure}[h]
\centerline{\includegraphics[width=\columnwidth]{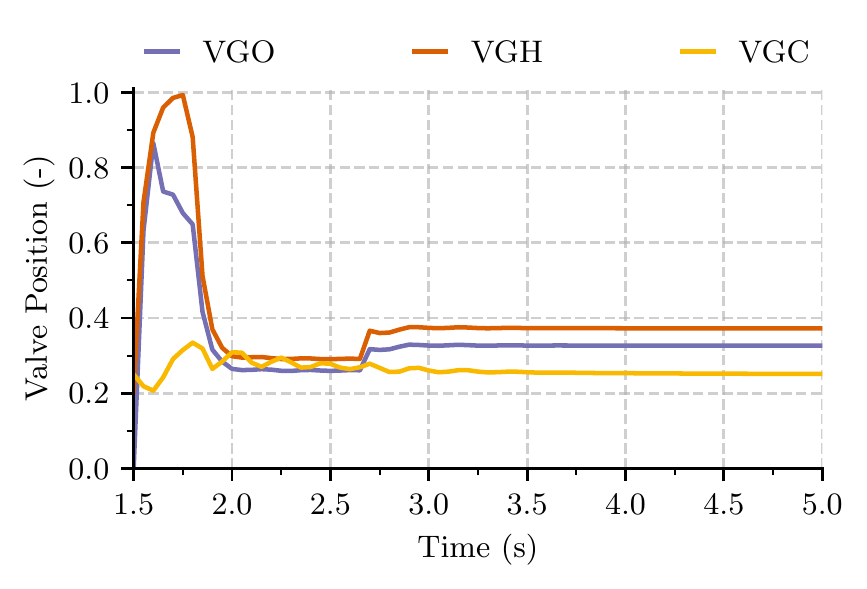}}
\caption{Manipulated valve positions by the RL agent for the \SI{80}{\bar} nominal start-up. The action clearly changes at $t=\SI{2.6}{\second}$, which is the time when the firing of turbine starter stops.}
\label{fig:rl_valve_positions_80}
\end{figure}

\begin{figure*}[p]
\centerline{\includegraphics[width=2\columnwidth]{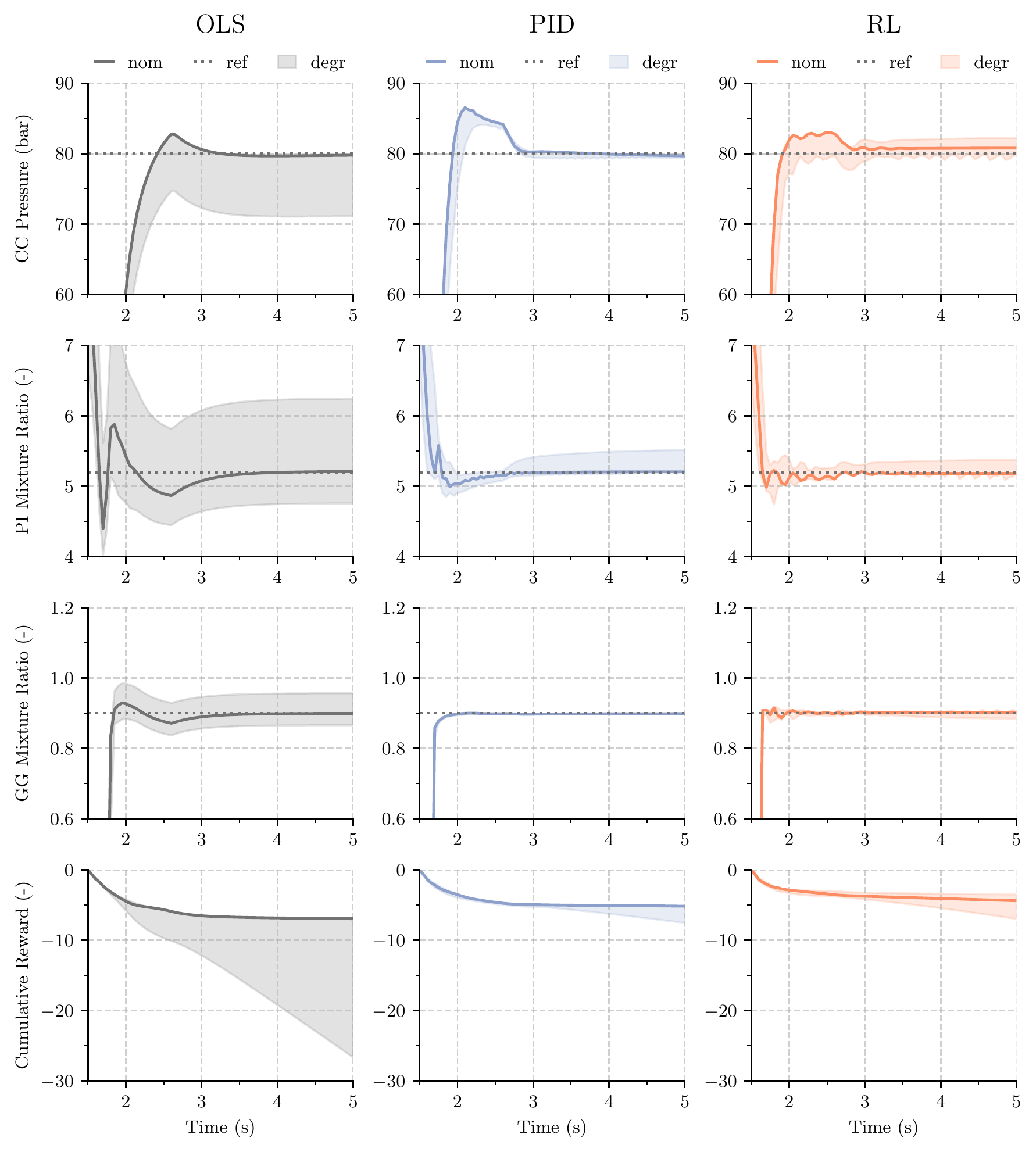}}
\caption{Comparison of the controlled variables for the \SI{80}{\bar} start-up. Shaded area marks the range of the controlled variable for different degraded efficiencies. At different turbine efficiencies the standard open-loop sequence provides significantly different steady-state values for the chamber pressure and the mixture ratios.}
\label{fig:comparison_80bar}
\end{figure*}

\begin{figure*}
\centerline{\includegraphics[width=2\columnwidth]{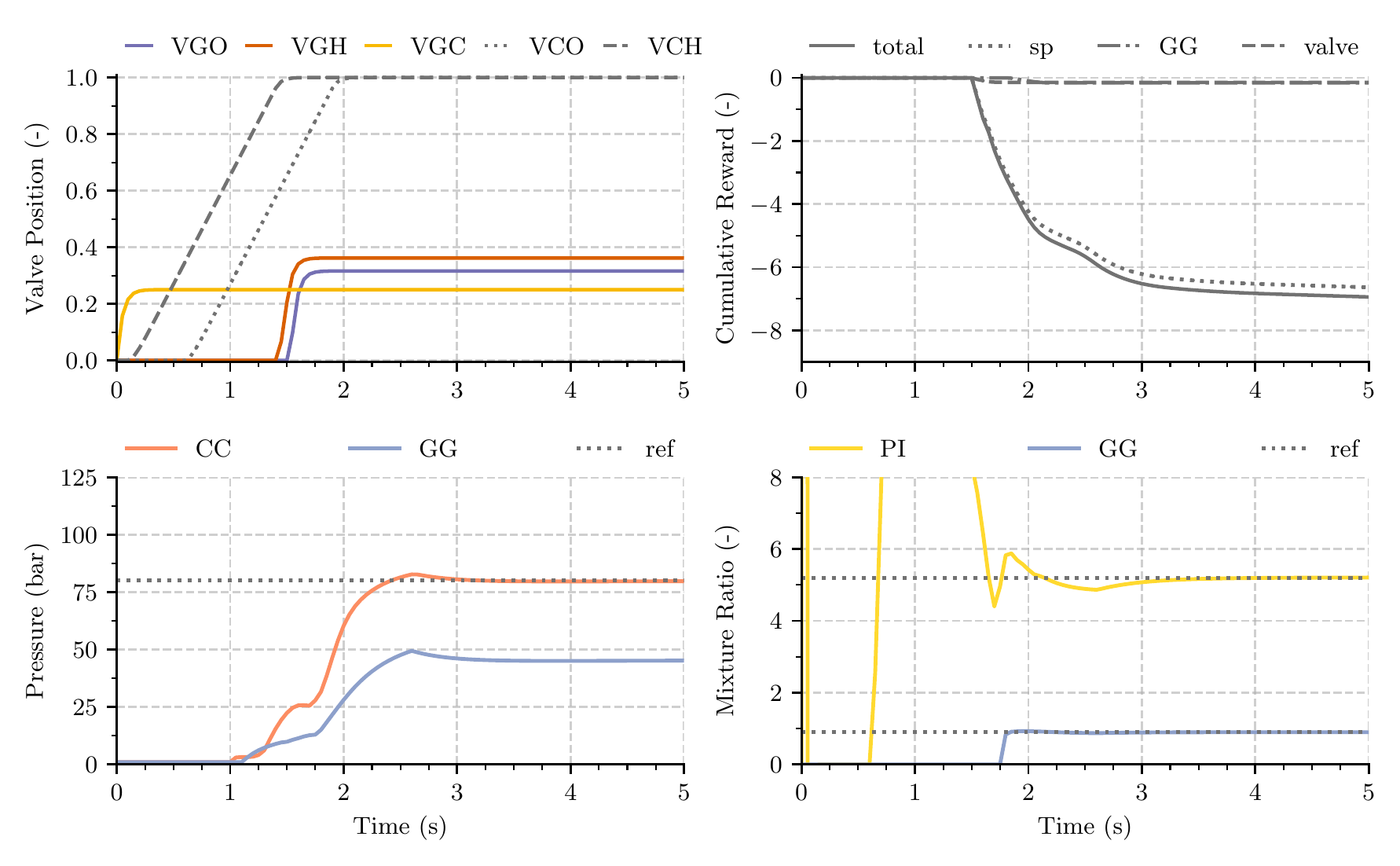}}
\caption{\SI{80}{\bar} nominal start-up sequence. The main combustion chamber pressure settles at \SI{80}{\bar}, while the gas-generator pressure reaches \SI{45}{\bar}. The reference mixture ratios are given by \num{5.2} and \num{0.9}. The engine reaches steady-state operating conditions after approximately \SI{4}{\second}. The cumulative reward for the OLS is dominated by the set-point error $r_{\text{sp}}$.}
\label{start_up80}
\end{figure*}

Fig. \ref{start_up80} shows the nominal OLS for a main combustion chamber pressure of \SI{80}{\bar}. The manipulated valve positions by the PID and RL agent for \SI{80}{\bar} are shown in Fig. \ref{fig:pid_valve_positions_80} and Fig. \ref{fig:rl_valve_positions_80}. Finally, Fig. \ref{fig:comparison_80bar} compares controller performances for different degraded turbine efficiencies.

\FloatBarrier
\bibliographystyle{bibliography/IEEEtran.bst}
\bibliography{bibliography/Publications.bib, bibliography/IEEEabrv.bib}

\begin{thebibliography}{10}
\providecommand{\url}[1]{#1}
\csname url@samestyle\endcsname
\providecommand{\newblock}{\relax}
\providecommand{\bibinfo}[2]{#2}
\providecommand{\BIBentrySTDinterwordspacing}{\spaceskip=0pt\relax}
\providecommand{\BIBentryALTinterwordstretchfactor}{4}
\providecommand{\BIBentryALTinterwordspacing}{\spaceskip=\fontdimen2\font plus
\BIBentryALTinterwordstretchfactor\fontdimen3\font minus
  \fontdimen4\font\relax}
\providecommand{\BIBforeignlanguage}[2]{{%
\expandafter\ifx\csname l@#1\endcsname\relax
\typeout{** WARNING: IEEEtran.bst: No hyphenation pattern has been}%
\typeout{** loaded for the language `#1'. Using the pattern for}%
\typeout{** the default language instead.}%
\else
\language=\csname l@#1\endcsname
\fi
#2}}
\providecommand{\BIBdecl}{\relax}
\BIBdecl

\bibitem{colas2019}
S.~Colas, S.~L. Gonidec, P.~Saunois, M.~Ganet, A.~Remy, and V.~Leboeuf,
  ``\BIBforeignlanguage{en}{A point of view about the control of a reusable
  engine cluster},'' in \emph{\BIBforeignlanguage{en}{Proceedings of the 8th
  {{European Conference}} for {{Aeronautics}} and {{Space Sciences}}}},
  {Madrid, Spain}, 2019.

\bibitem{meland1989}
L.~Meland and F.~Thompson, ``History of the {{Titan}} liquid rocket engines,''
  in \emph{25th {{Joint Propulsion Conference}}}.\hskip 1em plus 0.5em minus
  0.4em\relax {Sacramento, CA}: {American Institute of Aeronautics and
  Astronautics}, 1989.

\bibitem{lausten1985}
M.~Lausten, D.~Rousar, and S.~Buccella, ``\BIBforeignlanguage{en}{Carbon
  deposition with {{LOX}}/{{RP}}-1 propellants},'' in
  \emph{\BIBforeignlanguage{en}{21st {{Joint Propulsion Conference}}}}.\hskip
  1em plus 0.5em minus 0.4em\relax {Monterey,CA}: {American Institute of
  Aeronautics and Astronautics}, Jul. 1985.

\bibitem{bossard1989}
J.~A.~B. Bossard, ``Effect of {{Propellant Flowrate}} and {{Purity}} on
  {{Carbon Deposition}} in {{LO2}}/{{Methane Gas Generators}},'' May 1989.

\bibitem{roelke1973}
R.~J. Roelke, ``Miscellaneous losses. [tip clearance and disk friction],''
  {{NASA Technical Report}}, Jan. 1973.

\bibitem{hampson1984}
M.~E.~B. Hampson, ``Reusable rocket engine turbopump condition monitoring,'' in
  \emph{Space {{Systems Technology}}}, {Long Beach, CA}, 1984.

\bibitem{lorenzo1992}
C.~F. Lorenzo and J.~L. Musgrave, ``\BIBforeignlanguage{en}{Overview of rocket
  engine control},'' in \emph{\BIBforeignlanguage{en}{{{AIP Conference
  Proceedings}}}}, {Albuquerque, NM}, 1992.

\bibitem{chopinet2012}
J.-N. Chopinet, F.~Lassoudiere, G.~Roz, O.~Faye, S.~Le~Gonidec, P.~Alliot, and
  G.~Sylvain, ``\BIBforeignlanguage{en}{Progress of the development of an
  all-electric control system of a rocket engine},'' in
  \emph{\BIBforeignlanguage{en}{48th {{AIAA}}/{{ASME}}/{{SAE}}/{{ASEE Joint
  Propulsion Conference}} \& {{Exhibit}}}}, {Atlanta, Georgia}, Jul. 2012.

\bibitem{iannetti2017}
A.~Iannetti, ``\BIBforeignlanguage{en}{Prometheus, a {{LOX}}/{{LCH4}} reusable
  rocket engine},'' in \emph{\BIBforeignlanguage{en}{Proceedings of the 7th
  {{European Conference}} for {{Aeronautics}} and {{Space Sciences}}}},
  {Milano, Italy}, Jul. 2017.

\bibitem{asakawa2019}
H.~Asakawa, M.~Tanaka, T.~Takaki, and K.~Higashi,
  ``\BIBforeignlanguage{en}{Component tests of a {{LOX}}/methane full-expander
  cycle rocket engine: {{Electrically}} actuated valve},'' in
  \emph{\BIBforeignlanguage{en}{Proceedings of the 8th {{European Conference}}
  for {{Aeronautics}} and {{Space Sciences}}}}, 2019.

\bibitem{perez-roca2019}
S.~{P{\'e}rez-Roca}, J.~Marzat, H.~{Piet-Lahanier}, N.~Langlois, F.~Farago,
  M.~Galeotta, and S.~Le~Gonidec, ``\BIBforeignlanguage{en}{A survey of
  automatic control methods for liquid-propellant rocket engines},''
  \emph{\BIBforeignlanguage{en}{Progress in Aerospace Sciences}}, May 2019.

\bibitem{lopez2019}
V.~G. Lopez and F.~L. Lewis, ``\BIBforeignlanguage{en}{Dynamic {{Multiobjective
  Control}} for {{Continuous}}-{{Time Systems Using Reinforcement
  Learning}}},'' \emph{\BIBforeignlanguage{en}{IEEE Transactions on Automatic
  Control}}, Jul. 2019.

\bibitem{kiumarsi2018}
B.~Kiumarsi, K.~G. Vamvoudakis, H.~Modares, and F.~L. Lewis, ``Optimal and
  {{Autonomous Control Using Reinforcement Learning}}: {{A Survey}},''
  \emph{IEEE Transactions on Neural Networks and Learning Systems}, vol.~29,
  no.~6, pp. 2042--2062, Jun. 2018.

\bibitem{gu2016}
S.~Gu, E.~Holly, T.~Lillicrap, and S.~Levine, ``\BIBforeignlanguage{en}{Deep
  {{Reinforcement Learning}} for {{Robotic Manipulation}} with {{Asynchronous
  Off}}-{{Policy Updates}}},'' \emph{\BIBforeignlanguage{en}{arXiv:1610.00633
  [cs]}}, Nov. 2016.

\bibitem{yang2018}
Z.~Yang, K.~Merrick, L.~Jin, and H.~A. Abbass, ``Hierarchical {{Deep
  Reinforcement Learning}} for {{Continuous Action Control}},'' \emph{IEEE
  Transactions on Neural Networks and Learning Systems}, vol.~29, no.~11, pp.
  5174--5184, Nov. 2018.

\bibitem{mahmud2018}
M.~Mahmud, M.~S. Kaiser, A.~Hussain, and S.~Vassanelli, ``Applications of
  {{Deep Learning}} and {{Reinforcement Learning}} to {{Biological Data}},''
  \emph{IEEE Transactions on Neural Networks and Learning Systems}, vol.~29,
  no.~6, pp. 2063--2079, Jun. 2018.

\bibitem{heyer2020}
S.~Heyer, D.~Kroezen, and E.-J. Van~Kampen, ``\BIBforeignlanguage{en}{Online
  {{Adaptive Incremental Reinforcement Learning Flight Control}} for a
  {{CS}}-25 {{Class Aircraft}}},'' in \emph{\BIBforeignlanguage{en}{{{AIAA
  Scitech}} 2020 {{Forum}}}}, {Orlando, FL}, Jan. 2020.

\bibitem{gaudet2020}
B.~Gaudet, R.~Linares, and R.~Furfaro, ``\BIBforeignlanguage{en}{Deep
  reinforcement learning for six degree-of-freedom planetary landing},''
  \emph{\BIBforeignlanguage{en}{Advances in Space Research}}, vol.~65, no.~7,
  pp. 1723--1741, Apr. 2020.

\bibitem{spielberg2017}
S.~Spielberg, R.~Gopaluni, and P.~Loewen, ``Deep reinforcement learning
  approaches for process control,'' in \emph{2017 6th {{International
  Symposium}} on {{Advanced Control}} of {{Industrial Processes}}
  ({{AdCONIP}})}.\hskip 1em plus 0.5em minus 0.4em\relax {Taipei, Taiwan}:
  {IEEE}, May 2017, pp. 201--206.

\bibitem{musgrave1992a}
J.~L. Musgrave and D.~E. Paxson, ``A demonstration of an intelligent control
  system for a reusable rocket engine,'' {{NASA Technical Memorandum}} 105794,
  Jun. 1992.

\bibitem{nemeth1991}
E.~Nemeth, ``Reusable rocket engine intelligent control system framework
  design, phase 2,'' {{NASA Contractor Report}} 187213, Sep. 1991.

\bibitem{perez-roca2019b}
S.~{P{\'e}rez-Roca}, J.~Marzat, {\'E}.~Flayac, H.~{Piet-Lahanier}, N.~Langlois,
  F.~Farago, M.~Galeotta, and S.~L. Gonidec, ``\BIBforeignlanguage{en}{An {{MPC
  Approach}} to {{Transient Control}} of {{Liquid}}-{{Propellant Rocket
  Engines}}},'' \emph{\BIBforeignlanguage{en}{IFAC-PapersOnLine}}, vol.~52,
  no.~12, pp. 268--273, 2019.

\bibitem{perez-roca2018}
S.~{Perez-Roca}, N.~Langlois, J.~Marzat, H.~{Piet-Lahanier}, M.~Galeotta,
  F.~Farago, and S.~L. Gonidec, ``Derivation and {{Analysis}} of a
  {{State}}-{{Space Model}} for {{Transient Control}} of
  {{Liquid}}-{{Propellant Rocket Engines}},'' in \emph{2018 9th {{International
  Conference}} on {{Mechanical}} and {{Aerospace Engineering}} ({{ICMAE}})},
  {Budapest}, Jul. 2018.

\bibitem{sutton2018}
R.~S. Sutton and A.~G. Barto, \emph{\BIBforeignlanguage{en}{Reinforcement
  Learning: An Introduction}}, ser. Adaptive Computation and Machine Learning
  Series.\hskip 1em plus 0.5em minus 0.4em\relax {Cambridge, MA}: {The MIT
  Press}, 2018.

\bibitem{bertsekas2019}
D.~P. Bertsekas, \emph{\BIBforeignlanguage{English}{Reinforcement Learning and
  Optimal Control}}.\hskip 1em plus 0.5em minus 0.4em\relax {Athena
  Scientific}, 2019.

\bibitem{busoniu2018}
L.~Bu{\c s}oniu, T.~{de Bruin}, D.~Toli{\'c}, J.~Kober, and I.~Palunko,
  ``\BIBforeignlanguage{en}{Reinforcement learning for control:
  {{Performance}}, stability, and deep approximators},''
  \emph{\BIBforeignlanguage{en}{Annual Reviews in Control}}, vol.~46, pp.
  8--28, 2018.

\bibitem{lillicrap2019}
T.~P. Lillicrap, J.~J. Hunt, A.~Pritzel, N.~Heess, T.~Erez, Y.~Tassa,
  D.~Silver, and D.~Wierstra, ``\BIBforeignlanguage{en}{Continuous control with
  deep reinforcement learning},''
  \emph{\BIBforeignlanguage{en}{arXiv:1509.02971 [cs, stat]}}, Jul. 2019.

\bibitem{fujimoto2018}
S.~Fujimoto, H.~{van Hoof}, and D.~Meger, ``\BIBforeignlanguage{en}{Addressing
  {{Function Approximation Error}} in {{Actor}}-{{Critic Methods}}},''
  \emph{\BIBforeignlanguage{en}{arXiv:1802.09477}}, Oct. 2018.

\bibitem{wang2020}
X.~Wang, Y.~Gu, Y.~Cheng, A.~Liu, and C.~L.~P. Chen, ``Approximate
  {{Policy}}-{{Based Accelerated Deep Reinforcement Learning}},'' \emph{IEEE
  Transactions on Neural Networks and Learning Systems}, vol.~31, no.~6, pp.
  1820--1830, Jun. 2020.

\bibitem{haarnoja2017}
T.~Haarnoja, H.~Tang, P.~Abbeel, and S.~Levine,
  ``\BIBforeignlanguage{en}{Reinforcement {{Learning}} with {{Deep
  Energy}}-{{Based Policies}}},''
  \emph{\BIBforeignlanguage{en}{arXiv:1702.08165}}, Jul. 2017.

\bibitem{schulman2017a}
J.~Schulman, S.~Levine, P.~Moritz, M.~I. Jordan, and P.~Abbeel,
  ``\BIBforeignlanguage{en}{Trust {{Region Policy Optimization}}},''
  \emph{\BIBforeignlanguage{en}{arXiv:1502.05477 [cs]}}, Apr. 2017.

\bibitem{schulman2017}
J.~Schulman, F.~Wolski, P.~Dhariwal, A.~Radford, and O.~Klimov,
  ``\BIBforeignlanguage{en}{Proximal {{Policy Optimization Algorithms}}},''
  \emph{\BIBforeignlanguage{en}{arXiv:1707.06347 [cs]}}, Aug. 2017.

\bibitem{jin2018}
M.~Jin and J.~Lavaei, ``\BIBforeignlanguage{en}{Stability-certified
  reinforcement learning: {{A}} control-theoretic perspective},''
  \emph{\BIBforeignlanguage{en}{arXiv:1810.11505 [cs]}}, Oct. 2018.

\bibitem{vila2018}
J.~Vil{\'a}, J.~Moral, V.~{Fern{\'a}ndez-Villac{\'e}}, and J.~Steelant,
  ``\BIBforeignlanguage{en}{An {{Overview}} of the {{ESPSS Libraries}}:
  {{Latest Developments}} and {{Future}}},'' in
  \emph{\BIBforeignlanguage{en}{Space {{Propulsion}}}}, {Seville, Spain}, 2018.

\bibitem{stable-baselines}
A.~Hill, A.~Raffin, M.~Ernestus, A.~Gleave, A.~Kanervisto, R.~Traore,
  P.~Dhariwal, C.~Hesse, O.~Klimov, A.~Nichol, M.~Plappert, A.~Radford,
  J.~Schulman, S.~Sidor, and Y.~Wu, ``Stable baselines,'' \emph{GitHub
  repository}, 2018.

\bibitem{waxenegger-wilfing2020}
G.~{Waxenegger-Wilfing}, K.~Dresia, J.~C. Deeken, and M.~Oschwald,
  ``\BIBforeignlanguage{en}{Heat {{Transfer Prediction}} for {{Methane}} in
  {{Regenerative Cooling Channels}} with {{Neural Networks}}},''
  \emph{\BIBforeignlanguage{en}{Journal of Thermophysics and Heat Transfer}},
  vol.~34, no.~2, pp. 347--357, Apr. 2020.

\bibitem{dresia2019}
K.~Dresia, G.~{Waxenegger-Wilfing}, J.~Riccius, J.~Deeken, and M.~Oschwald,
  ``\BIBforeignlanguage{en}{Numerically {{Efficient Fatigue Life Prediction}}
  of {{Rocket Combustion Chambers}} using {{Artificial Neural Networks}}},'' in
  \emph{\BIBforeignlanguage{en}{Proceedings of the 8th {{European Conference}}
  for {{Aeronautics}} and {{Space Sciences}}}}.\hskip 1em plus 0.5em minus
  0.4em\relax {Madrid, Spain}: {Proceedings of the 8th European Conference for
  Aeronautics and Space Sciences. Madrid, Spain}, 2019.

\bibitem{iffly1999}
A.~Iffly and M.~Brixhe, ``\BIBforeignlanguage{en}{Performance model of the
  {{Vulcain Ariane}} 5 main engine},'' in \emph{\BIBforeignlanguage{en}{35th
  {{Joint Propulsion Conference}} and {{Exhibit}}}}.\hskip 1em plus 0.5em minus
  0.4em\relax {Los Angeles,CA}: {American Institute of Aeronautics and
  Astronautics}, Jun. 1999.

\bibitem{ryan1986}
R.~Ryan and L.~Gross, ``Effects of geometry and materials on low cycle fatigue
  life of turbine blades in {{LOX}}/hydrogen rocket engines,'' in \emph{22nd
  {{Joint Propulsion Conference}}}.\hskip 1em plus 0.5em minus 0.4em\relax
  {American Institute of Aeronautics and Astronautics}, 1986.

\bibitem{pempie2001}
P.~Pempie, T.~Froehlich, and H.~Vernin,
  ``\BIBforeignlanguage{en}{{{LOX}}/methane and {{LOX}}/kerosene high thrust
  engine trade-off},'' in \emph{\BIBforeignlanguage{en}{37th {{Joint Propulsion
  Conference}} and {{Exhibit}}}}.\hskip 1em plus 0.5em minus 0.4em\relax {Salt
  Lake City,UT}: {American Institute of Aeronautics and Astronautics}, Jul.
  2001.

\bibitem{cominos2002}
P.~Cominos and N.~Munro, ``\BIBforeignlanguage{en}{{{PID}} controllers: Recent
  tuning methods and design to specification},''
  \emph{\BIBforeignlanguage{en}{IEE Proceedings - Control Theory and
  Applications}}, vol. 149, no.~1, pp. 46--53, Jan. 2002.

\bibitem{DEAP_JMLR2012}
F.-A. Fortin, F.-M. De~Rainville, M.-A. Gardner, M.~Parizeau, and {Christian
  Gagn\'e}, ``{{DEAP}}: {{Evolutionary}} algorithms made easy,'' \emph{Journal
  of Machine Learning Research}, vol.~13, pp. 2171--2175, Jul. 2012.

\bibitem{tobin2017}
J.~Tobin, R.~Fong, A.~Ray, J.~Schneider, W.~Zaremba, and P.~Abbeel, ``Domain
  randomization for transferring deep neural networks from simulation to the
  real world,'' in \emph{2017 {{IEEE}}/{{RSJ International Conference}} on
  {{Intelligent Robots}} and {{Systems}} ({{IROS}})}, Sep. 2017, pp. 23--30.

\bibitem{achiam2017}
J.~Achiam, D.~Held, A.~Tamar, and P.~Abbeel,
  ``\BIBforeignlanguage{en}{Constrained {{Policy Optimization}}},''
  \emph{\BIBforeignlanguage{en}{arXiv:1705.10528 [cs]}}, May 2017.

\bibitem{ray1994}
A.~Ray, X.~Dai, M.-K. Wu, M.~Carpino, and C.~F. Lorenzo,
  ``\BIBforeignlanguage{en}{Damage-mitigating control of a reusable rocket
  engine},'' \emph{\BIBforeignlanguage{en}{Journal of Propulsion and Power}},
  vol.~10, no.~2, pp. 225--234, Mar. 1994.

\bibitem{optuna_2019}
T.~Akiba, S.~Sano, T.~Yanase, T.~Ohta, and M.~Koyama, ``Optuna: {{A}}
  next-generation hyperparameter optimization framework,'' in \emph{Proceedings
  of the 25rd {{ACM SIGKDD}} International Conference on Knowledge Discovery
  and Data Mining}, 2019.

\bibitem{uhlenbeck1930}
G.~E. Uhlenbeck and L.~S. Ornstein, ``On the {{Theory}} of the {{Brownian
  Motion}},'' \emph{Physical Review}, vol.~36, no.~5, pp. 823--841, Sep. 1930.

\end{thebibliography}




\begin{IEEEbiography}[{\includegraphics[width=1in,height=1.25in,clip,keepaspectratio]{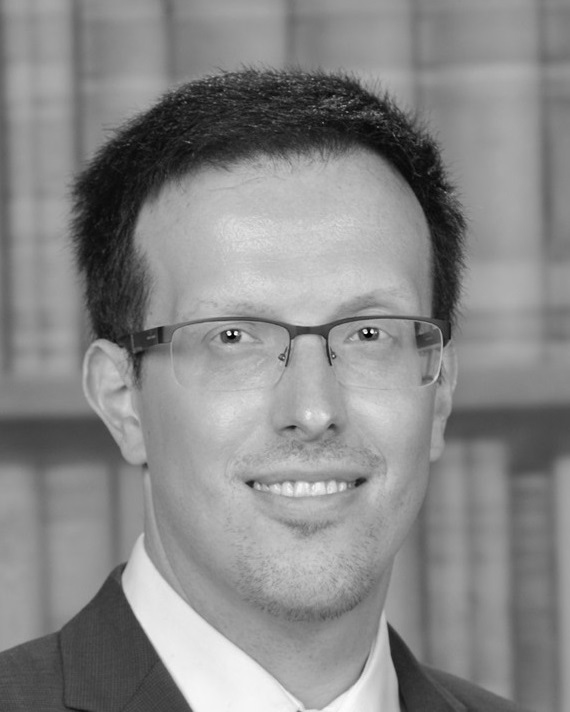}}]{G\"unther Waxenegger-Wilfing} received his Ph.D. degree in theoretical physics from the University of Vienna. He is a senior research scientist for intelligent engine control at the German Aerospace Center (DLR) Institute of Space Propulsion and a lecturer at the University of W\"urzburg. As part of the DLR project AMADEUS, he manages the investigation of the use of artificial intelligence in space transportation. He previously was the lead quantitative analyst of Nova Portfolio Verm\"ogensManagement, where he focused on applied machine learning, time series analysis, and forecasting. G\"unther Waxenegger-Wilfing’s research interests include deep learning for control and condition monitoring, optimal control, model-free as well as model-based reinforcement learning, and applications in autonomous launch vehicles and landers.
\end{IEEEbiography}

\begin{IEEEbiography}[{\includegraphics[width=1in,height=1.25in,clip,keepaspectratio]{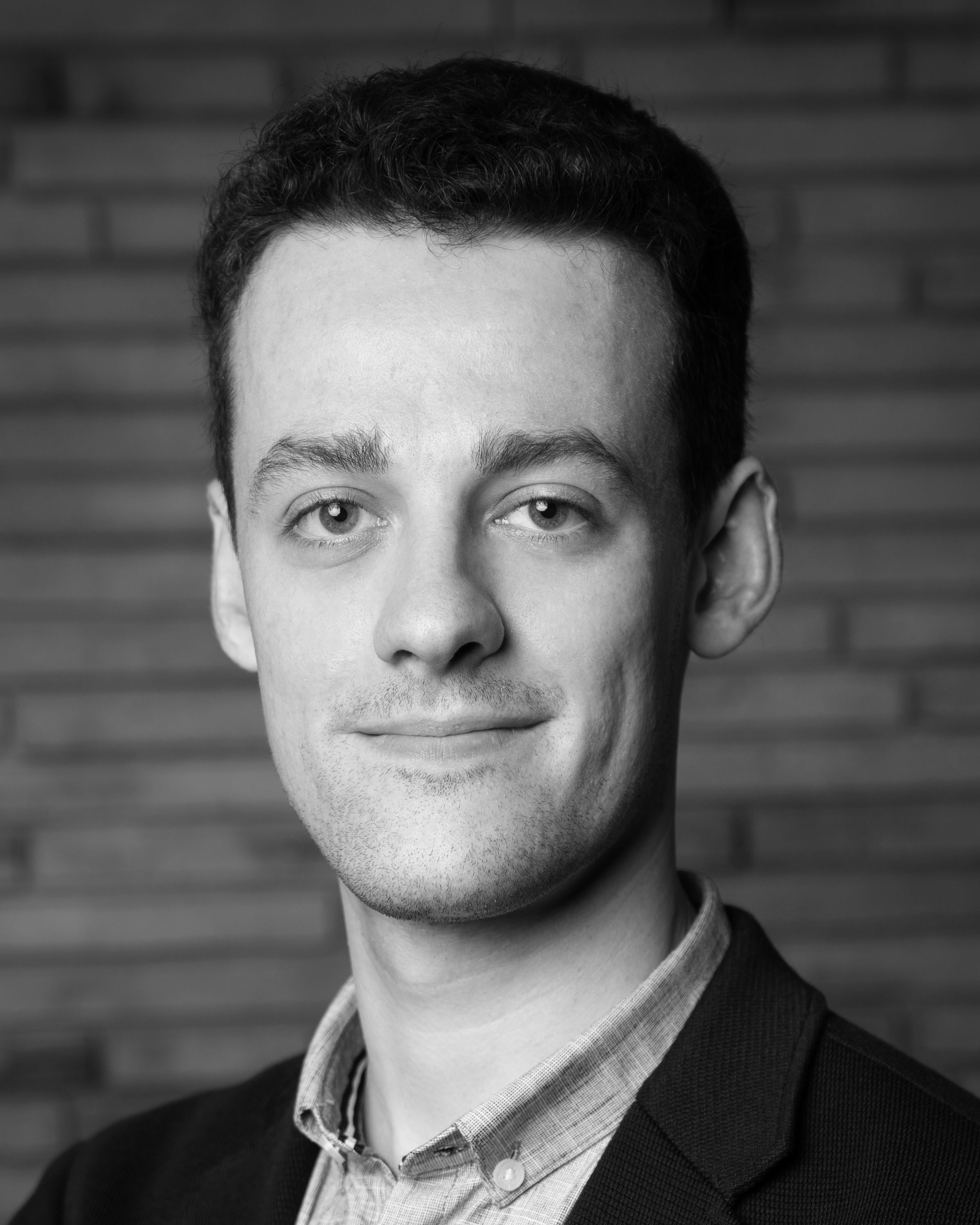}}]{Kai Dresia} received his Masters degree in aerospace engineering from RWTH Aachen University. In his master thesis, he investigated the application of artificial neural networks for heat transfer modeling into supercritical methane flowing in cooling channels of a regeneratively cooled combustion chamber. He is currently a Ph.D. candidate at the German Aerospace Center (DLR) Institute of Space Propulsion, where his work focuses on combining machine learning with physics-based modeling for rocket engine control and condition monitoring.
\end{IEEEbiography}

\begin{IEEEbiography}[{\includegraphics[width=1in,height=1.25in,clip,keepaspectratio]{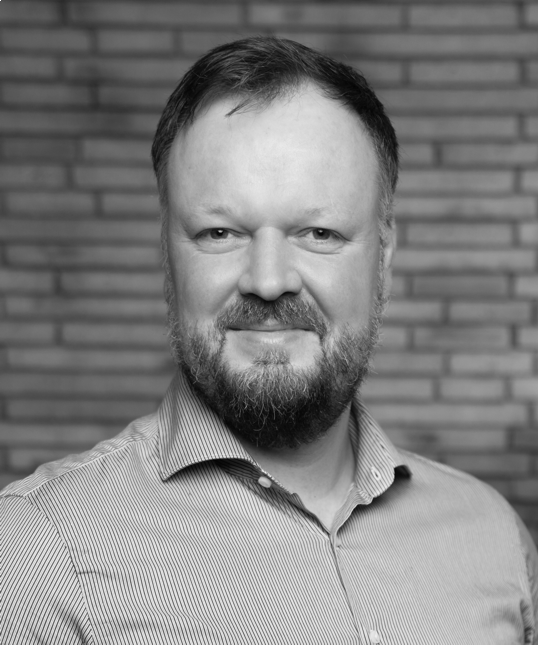}}]{Jan Deeken} received his Ph.D. in aerospace engineering from the university of Stuttgart for his work on a novel injection concept for gas-liquid injection in high pressure rocket combustion chambers. He is currently head of the rocket engine system analysis group within the department of rocket propulsion at the DLR Institute of Space Propulsion in Lampoldshausen, Germany. He is also managing the DLR LUMEN project, which aims at developing and operating a LOX/LNG expander-bleed engine testbed in the 25kN thrust class.
\end{IEEEbiography}

\begin{IEEEbiography}[{\includegraphics[width=1in,height=1.25in,clip,keepaspectratio]{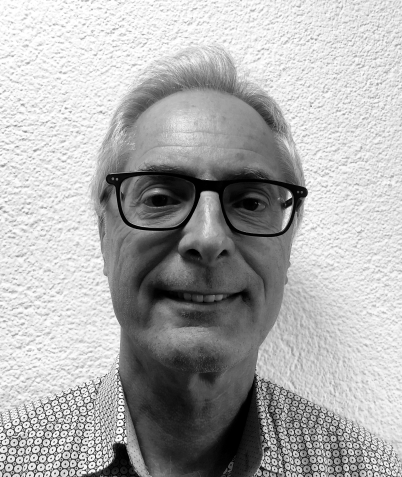}}]{Michael Oschwald} is a professor for Space Propulsion of the RWTH Aachen University and at the same time head of the Department of Rocket Propulsion at the Institute of Space Propulsion at the German Aerospace Center (DLR). His fields of activity cover all aspects of rocket engine design and operation with a specific focus on cryogenic propulsion. His department at DLR has a long heritage in experimental investigations of high-pressure combustion, heat transfer, combustion instabilities, and expansion nozzles. In parallel to the experimental work, numerical tools are developed in his department to predict the behavior of rocket engines and their components.
\end{IEEEbiography}

\end{document}